\documentclass{article}

\usepackage{iclr2027_conference,times}

\usepackage{iftex}
\ifXeTeX
  \usepackage[T1]{fontenc}
\fi

\usepackage{amsmath,amsfonts,bm}

\def\eqref#1{equation~\ref{#1}}

\def\1{\bm{1}}

\DeclareMathAlphabet{\mathsfit}{\encodingdefault}{\sfdefault}{m}{sl}
\SetMathAlphabet{\mathsfit}{bold}{\encodingdefault}{\sfdefault}{bx}{n}

\newcommand{\E}{\mathbb{E}}

\newcommand{\R}{\mathbb{R}}

\usepackage{amsmath,amssymb,mathtools}
\usepackage{amsthm}
\usepackage{booktabs,multirow,array}
\usepackage{graphicx}
\usepackage{float}
\usepackage{capt-of}
\usepackage{placeins}
\usepackage{microtype}
\usepackage{hyperref}
\usepackage{url}
\usepackage{titletoc}

\newtheorem{proposition}{Proposition}

\hypersetup{hidelinks}

\newcommand{\method}{\textnormal{QuantMLA}}

\providecommand{\E}{\mathbb{E}}
\providecommand{\R}{\mathbb{R}}

\title{QuantMLA: Function-Aligned Dual-Path Quantization for Low-Bit MLA KV Caching}

\iclrfinalcopy

\providecommand{\ICLRauthorrow}[1]{}
\renewcommand{\ICLRauthorrow}[1]{%
  \makebox[\dimexpr\textwidth-2\tabcolsep\relax][l]{%
    \normalfont\normalsize #1%
  }%
}

\author{%
  \ICLRauthorrow{%
    \textbf{Zunhai Su}\textsuperscript{1,2}%
    \thanks{Zunhai Su and Yuxuan Sun contributed equally to this work.}%
    \hspace{0.5em}%
    \textbf{Yuxuan Sun}\textsuperscript{2}\footnotemark[1]%
    \hspace{0.5em}%
    \textbf{Jianchao Tan}\textsuperscript{2}%
    \hspace{0.5em}%
    \textbf{Tao Zhang}\textsuperscript{3}%
    \hspace{0.5em}%
    \textbf{Ruihan Hu}\textsuperscript{4}%
    \hspace{0.5em}%
    \textbf{Yuchen Xie}\textsuperscript{2}%
  }\\[1pt]
  \ICLRauthorrow{%
    \textbf{Xunliang Cai}\textsuperscript{2}%
    \hspace{0.5em}%
    \textbf{Ngai Wong}\textsuperscript{1}%
  }\\[3pt]
  \ICLRauthorrow{%
    \small
    \textsuperscript{1}The University of Hong Kong\hspace{0.8em}%
    \textsuperscript{2}Meituan LongCat Team\hspace{0.8em}%
    \textsuperscript{3}South China University of Technology%
  }\\[1pt]
  \ICLRauthorrow{%
    \small
    \textsuperscript{4}Harbin Institute of Technology%
  }%
}

\begin{document}

\maketitle

\begin{abstract}
Multi-Head Latent Attention (MLA) enables expressive multi-head attention with compact caches for its content and decoupled RoPE paths, yet cache memory still scales linearly with context length and batch size.
Existing methods primarily quantize the content cache to FP8 while retaining the RoPE key cache at high precision.
Low-bit RoPE quantization remains poorly understood, leaving joint low-bit compression of the content and RoPE caches largely unexplored.
In this work, we establish a systematic model of MLA's dual-path quantization errors, characterizing their distinct effects on attention-output distortion and explaining the pronounced amplification of RoPE-path errors.
Guided by this analysis, we introduce \method{}, a function-aligned framework for low-bit dual-path quantization.
We derive path-specific transformation spaces that preserve full-precision computation while remaining fully fusible into model parameters offline, eliminating online transformation overhead.
Within these spaces, \method{} learns path-specific transformations with function-aligned objectives: attention-output reconstruction captures the content path's coupled matching and aggregation errors, while positional QK reconstruction preserves the RoPE-induced component of the attention logits and admits a theoretical bound on output distortion.
Across four MLA model families, \method{} enables, to our knowledge, the first reported joint INT4 caching of the content and RoPE caches with minimal accuracy degradation.
Further compressing the content cache to INT2 while retaining the RoPE key cache at INT4 maintains competitive performance on challenging reasoning and code benchmarks.
We develop a native low-bit MLA attention kernel that integrates unpacking and dequantization directly into attention computation.
The physical cache layout provides $3.59\times$ compression at 128K context, while a cache-pressure serving workload achieves $5.168\times$ higher whole-job output throughput than BF16.
The code will be released upon acceptance.

\end{abstract}

\section{Introduction}
\label{sec:introduction}

\begin{figure}[!t]
    \centering
    \vspace{-5mm}
    \includegraphics[width=\textwidth]{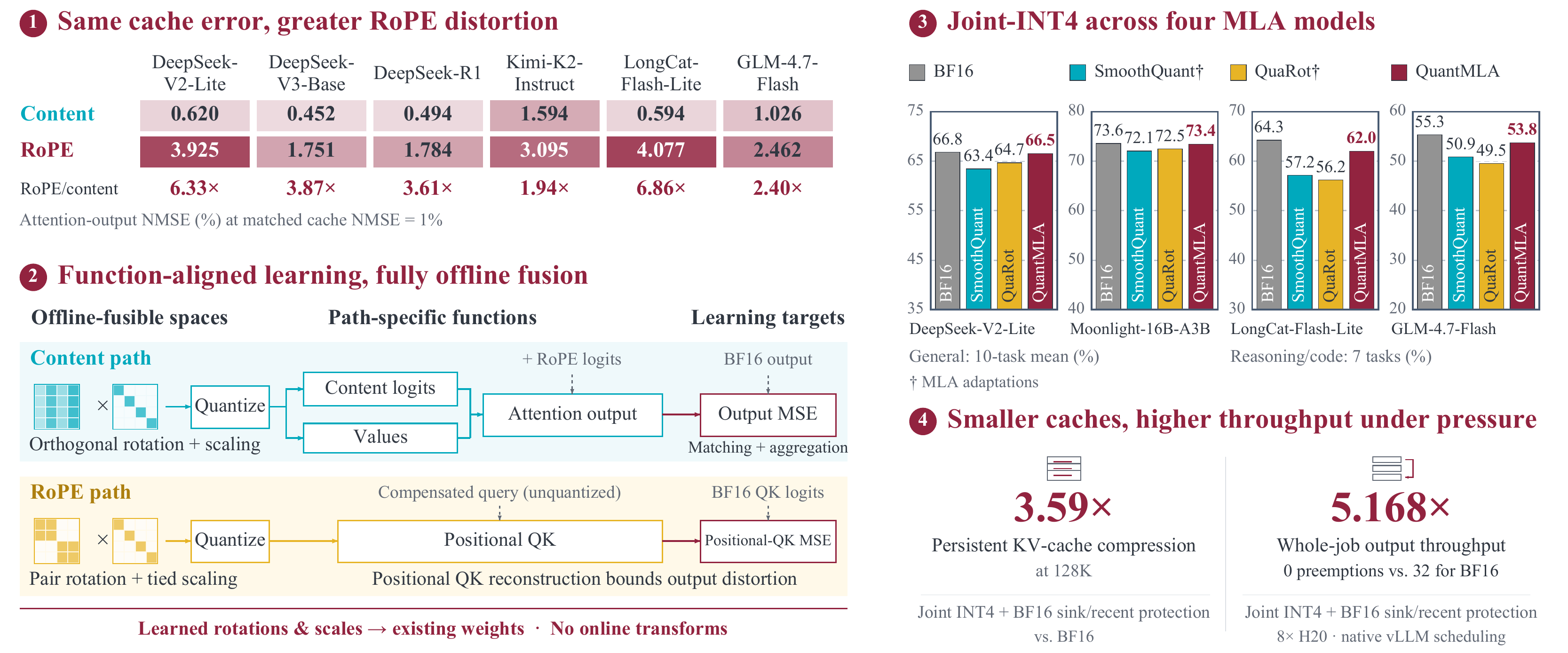}
    \caption{\textbf{Overview of QuantMLA.}
    Our analysis reveals pronounced RoPE-path error amplification under matched cache reconstruction error.
    Guided by this analysis, we introduce \method{}, which learns path-specific transformations with function-aligned objectives and fuses them entirely offline for accurate and efficient low-bit MLA inference.}
    \label{fig:quantmla_teaser}
\end{figure}

Attention enables Transformer-based large language models (LLMs) to integrate information across long contexts, but autoregressive decoding relies on a key--value (KV) cache whose storage and access incur substantial memory and bandwidth costs.
This bottleneck has motivated efficient attention mechanisms and architectures, including linear attention~\citep{linear_attention}, sparse attention~\citep{nsa}, and hybrid architectures~\citep{hla,kimi_linear}.
Multi-Head Latent Attention (MLA) reconciles expressive multi-head attention with compact KV storage through two cached paths: a shared low-rank content latent from which keys and values are derived, and a decoupled RoPE key cache~\citep{deepseekv2}.
This design has been adopted by leading models including DeepSeek-V3, Kimi K3, LongCat-Flash, LongCat-2.0, and GLM-5.3~\citep{deepseekv3,kimi_k3,longcat_report,longcat2,glm53}.

Despite its compact representation, MLA cache memory still scales linearly with context length and serving batch size, motivating further compression through low-bit KV-cache quantization~\citep{kivi,rotatekv,oscar}.
Existing MLA quantization approaches primarily compress the content cache to FP8 while retaining the RoPE key cache at high precision~\citep{flashmla,snapmla}.
For example, FlashMLA stores the DeepSeek-V3.2 content cache in FP8 and the RoPE key cache in BF16, while SnapMLA co-designs content-cache quantization and execution but likewise preserves high-precision RoPE keys.
The functional impact of RoPE-path quantization therefore remains poorly understood, leaving joint low-bit compression largely unexplored.
Appendix~\ref{app:related_work} reviews transformation-based quantization and MLA KV-cache quantization in detail.

In this work, we systematically analyze MLA's dual-path quantization through cross-model empirical characterization, path-wise functional decomposition, and mechanistic modeling.
Across six models from four families, spanning 16B to 1T total parameters, RoPE-path quantization consistently induces greater attention-output distortion than content-path quantization under matched cache reconstruction error.
We show that content-path errors propagate through attention matching, value aggregation, and their interaction, whereas RoPE-path errors act exclusively through the RoPE-induced component of the attention logits.
An operator--error model further explains this amplification through attention sensitivity, quantization-error energy allocation, and directional coupling.
These results show that cache reconstruction error alone is an inadequate proxy for functional distortion.

Guided by this analysis, we introduce \method{}, a function-aligned framework for low-bit dual-path quantization.
We derive path-specific offline-fusible transformation spaces and optimize within them using function-aligned objectives tailored to the two paths' distinct functional error routes: attention-output reconstruction for the content path and positional QK reconstruction for the RoPE path, the latter admitting a theoretical bound on output distortion.
Thus, each path's structure determines its offline-fusible transformation space, while its functional error route determines the corresponding learning objective.

We develop a carefully engineered native low-bit MLA attention kernel that integrates unpacking and dequantization directly into attention computation.
Across four MLA model families, we evaluate downstream accuracy together with cache footprint, native attention execution, and end-to-end vLLM serving behavior.
Figure~\ref{fig:quantmla_teaser} summarizes QuantMLA and its main findings.
Our main contributions are summarized as follows:

\begin{itemize}
    \item \textbf{A systematic analysis of MLA dual-path quantization.}
    We characterize the distinct functional error routes of the content and RoPE paths and develop an operator--error model that explains RoPE-path error amplification, validated through cross-model analyses and controlled mechanistic interventions.

    \item \textbf{Function-aligned dual-path quantization.}
    We introduce \method{}, which derives path-specific offline-fusible transformation spaces for MLA's dual paths and optimizes within them using function-aligned objectives: attention-output reconstruction for the content path and positional QK reconstruction for the RoPE path.

    \item \textbf{Accurate joint low-bit MLA caching.}
    Across four MLA families, \method{} enables, to our knowledge, the first reported joint INT4 caching of the content and RoPE caches with minimal accuracy degradation.
    Further compressing the content cache to INT2 while retaining INT4 RoPE keys maintains competitive performance on challenging reasoning and code benchmarks.

    \item \textbf{Memory-efficient native MLA execution.} We develop a native low-bit MLA backend that integrates unpacking and dequantization into attention. System evaluations demonstrate up to $3.59\times$ KV-cache compression at 128K and $5.168\times$ BF16's whole-job output throughput in the evaluated cache-pressure workload under the same native vLLM scheduling policy.
\end{itemize}
\section{Modeling Dual-Path Quantization Error}
\label{sec:functional_risk}

Our analysis progresses from MLA's dual-path computation to empirical observation and mechanistic explanation.
We first formulate the content and RoPE cache paths (Section~\ref{subsec:mla_formulation}), then reveal unequal output distortion under matched cache reconstruction error (Section~\ref{subsec:matched_error}), and finally explain it through functional error routes, attention sensitivity, and quantization-error geometry (Section~\ref{subsec:error_propagation}).

\subsection{MLA Dual-Path Cache Formulation}
\label{subsec:mla_formulation}

Following MLA's shared-latent and decoupled-RoPE formulation~\citep{deepseekv2,deepseekv3}, let $C\in\R^{n\times d_c}$ denote the cached content latent and $K_P\in\R^{n\times d_r}$ the decoupled RoPE key.
Conceptually, per-head content keys and values are reconstructed from $C$; at inference, MLA absorbs the up-projections into the query and output projections, avoiding explicit reconstruction~\citep{deepseekv2,snapmla}.
Using row-vector notation, define the absorbed content query $\bar Q_{C,h}=Q_{C,h}W_{K,h}^{\top}$ and value--output map $B_h=W_{V,h}W_{O,h}$ for head $h$.
Then
\begin{equation}
\begin{aligned}
S_h
&=\tau\!\left(
\bar Q_{C,h}C^\top
+
Q_{P,h}K_P^\top
\right)+M_{\mathrm{attn}},\\
A_h
&=\operatorname{softmax}(S_h),
\qquad
Y=\sum_h A_hCB_h .
\end{aligned}
\label{eq:mla_compact}
\end{equation}
where $Q_{P,h}$ is the RoPE-encoded query component, $M_{\mathrm{attn}}$ the additive causal mask, and $\tau$ the model-specific attention scale.
Only $\{C,K_P\}$ is cached, requiring $d_c+d_r$ elements per token and layer.
The content latent contributes to both matching through $\bar Q_{C,h}C^\top$ and aggregation through $A_hCB_h$, whereas the RoPE key contributes only to positional matching through $Q_{P,h}K_P^\top$.
We refer to these computations as the \emph{content path} and \emph{RoPE path}, respectively.

\subsection{Matched Cache Error, Unequal Output Distortion}
\label{subsec:matched_error}

\begin{figure}[t]
    \centering
    \includegraphics[width=\textwidth]{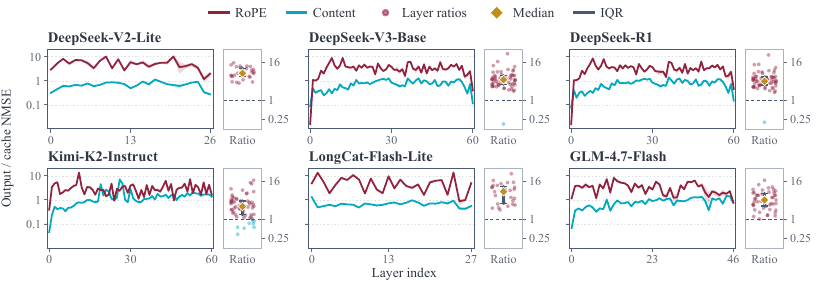}
    \caption{\textbf{Layer-wise functional response under matched cache reconstruction error.}
    Curves show the content- and RoPE-path responses $R_b$ across layers for six MLA models.
    The adjacent distributions show the corresponding layer-wise RoPE/content response ratios; diamonds and whiskers denote medians and interquartile ranges, and dashed lines mark equal response.}
    \label{fig:layerwise_error_amplification}
\end{figure}

We begin with an empirical comparison of the content and RoPE paths under matched cache reconstruction error.
We perturb one path's cache at a time using error directions induced by INT4 quantization and rescale each perturbation to a common cache NMSE of $0.01$.
Our study covers six models from four families, spanning 16B--1T total parameters: DeepSeek-V2-Lite, DeepSeek-V3-Base, DeepSeek-R1, Kimi-K2-Instruct, LongCat-Flash-Lite, and GLM-4.7-Flash.
To compare functional distortion across layers, we define the path-wise response as $R_b=\operatorname{NMSE}_{\mathrm{out},b}/\operatorname{NMSE}_{\mathrm{cache},b}$ for $b\in\{C,P\}$, measuring output distortion per unit cache reconstruction error.
Across the six models, the ratio of model-level mean responses between the RoPE and content paths ranges from $1.941\times$ to $6.864\times$.
Figure~\ref{fig:layerwise_error_amplification} further reveals substantial layer-wise variation in both responses and their RoPE-to-content ratios.
These results show that matched cache reconstruction error can yield markedly different functional distortion, despite the substantially smaller RoPE key cache.
Appendix~\ref{app:matched_error} provides the evaluation protocol, aggregation procedure, and detailed per-model and per-layer results.

\subsection{Functional Error Routes and Amplification}
\label{subsec:error_propagation}

The matched-error discrepancy raises two questions: how do errors from the two cache paths reach the attention output, and what determines their amplification?
We first derive their exact finite error routes, then model local amplification through attention sensitivity and quantization-error geometry.

\paragraph{Finite error routes.}
Let $\Delta C$ and $\Delta K_P$ denote quantization errors in the content cache and RoPE key cache.
Under the absorbed formulation, their score perturbations for head $h$ are
\begin{equation}
    \Delta S_{C,h}
    =\tau\bar Q_{C,h}\Delta C^\top,
    \qquad
    \Delta S_{P,h}
    =\tau Q_{P,h}\Delta K_P^\top.
    \label{eq:path_score_perturbations}
\end{equation}
Let $\Delta A_{b,h}=\operatorname{softmax}(S_h+\Delta S_{b,h})-A_h$ denote the corresponding attention change for path $b\in\{C,P\}$.
Perturbing one path at a time gives the exact finite output changes
\begin{equation}
\Delta Y_C=\sum_h\!\left(
\Delta A_{C,h}CB_h
+A_h\Delta C B_h
+\Delta A_{C,h}\Delta C B_h
\right),
\qquad
\Delta Y_P=\sum_h\Delta A_{P,h}CB_h.
\label{eq:finite_path_errors}
\end{equation}
Content-path errors therefore affect both \emph{attention matching} and \emph{value aggregation}, including their interaction, whereas RoPE-path errors affect only attention matching through the positional logits.

\paragraph{Local amplification model.}
Vectorize the stored state of path $b\in\{C,P\}$ as $c_b\in\R^{D_b}$, its quantization error as $e_b$, and the post-output-projection attention output as $y=\operatorname{vec}(Y)$.
Let $J_b=\partial y/\partial c_b$ denote the Jacobian from local cache perturbations to output perturbations, and define $H_b=J_b^\top J_b$.
We quantify local amplification along the quantization-error direction by
\begin{equation}
    G_b(e_b)
    =\frac{\|J_be_b\|_2^2}{\|e_b\|_2^2}
    =\frac{e_b^\top H_be_b}{e_b^\top e_b}.
    \label{eq:functional_gain}
\end{equation}
This Rayleigh quotient measures local functional sensitivity, so equal-energy errors can induce markedly different output distortion.
Locally, $\operatorname{NMSE}_{\mathrm{out},b}\approx\operatorname{NMSE}_{\mathrm{cache},b}(\|c_b\|_2^2/\|y\|_2^2)G_b(e_b)$, separating cache-error magnitude from functional amplification.

To expose the sources of this gain, expand $e_b^\top H_be_b$ into its diagonal contribution $\sum_i H_{b,ii}e_{b,i}^2$ and off-diagonal contribution $\sum_{i\neq j}H_{b,ij}e_{b,i}e_{b,j}$.
With $T_b=\operatorname{tr}(H_b)/D_b>0$, define
\[
A_b=\frac{\sum_i H_{b,ii}e_{b,i}^2}{T_b\|e_b\|_2^2},
\qquad
O_b=\frac{\sum_{i\neq j}H_{b,ij}e_{b,i}e_{b,j}}{\|e_b\|_2^2}.
\]
The gain then admits the exact decomposition
\begin{equation}
    \boxed{G_b=T_bA_b+O_b.}
    \label{eq:gain_decomposition}
\end{equation}
Here, $T_b$ measures average operator sensitivity, $A_b$ captures the allocation of quantization-error energy over sensitive coordinates, and the signed term $O_b$ captures cross-coordinate directional coupling.
Thus, equal-energy errors can differ in functional impact because of both operator sensitivity and error geometry.
A RoPE-path error aligned with sensitive directions can therefore outweigh the content path's coupled errors despite entering through only one functional route.
Appendix~\ref{app:gain_derivation} gives the full derivation and joint-path extension, while Appendix~\ref{app:functional_analysis} describes operator estimation and validates the model through cross-layer prediction and controlled equal-energy interventions.
\section{QuantMLA: Function-Aligned Dual-Path Quantization}
\label{sec:quantmla}

\begin{figure}[t]
    \centering
    \includegraphics[width=\textwidth]{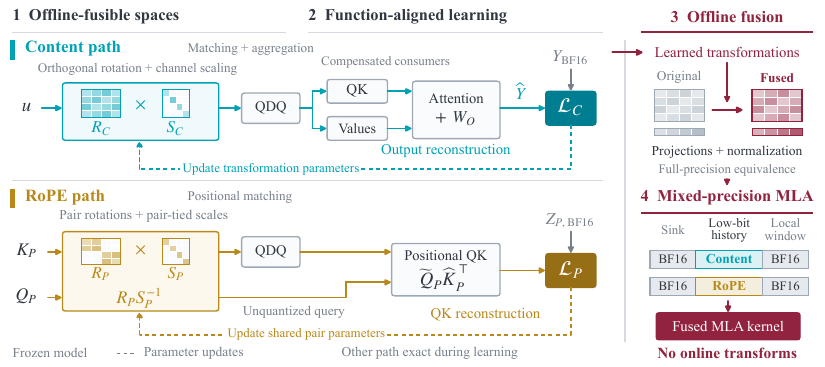}
    \caption{\textbf{QuantMLA learning and execution pipeline.}
    Content and RoPE transformations are optimized with attention-output and positional QK reconstruction, respectively, and fused offline.
    The native low-bit MLA backend stores packed cache representations while serving protected sink and recent tokens from BF16 buffers.
    QDQ denotes simulated quantize--dequantize.}
    \label{fig:quantmla_pipeline}
\end{figure}

Building on the analysis in Section~\ref{sec:functional_risk}, we introduce \method{}, which translates MLA's dual-path asymmetry into path-specific offline-fusible transformation spaces and function-aligned learning objectives.
We first derive the offline-fusible transformations for each path (Section~\ref{subsec:transformation_spaces}), then optimize them according to their functional error routes (Section~\ref{subsec:learning_objectives}), and finally deploy the fused model with a native low-bit MLA backend (Section~\ref{subsec:kernel_design}; Figure~\ref{fig:quantmla_pipeline}).

\subsection{Path-Specific Offline-Fusible Transformation Spaces}
\label{subsec:transformation_spaces}

The two cache paths impose distinct constraints on equivalent reparameterization: the shared content latent participates in both matching and aggregation, whereas the decoupled RoPE key must additionally respect RoPE's pairwise rotational structure.

\paragraph{Content path.}
Let $c=u\Gamma$ denote the content latent, where $u=\operatorname{RMSNorm}_1(x)$ is the weightless normalized coordinate~\citep{rmsnorm} and $\Gamma=\operatorname{Diag}(\gamma)$ contains the original normalization weights.
We reparameterize the cache with an orthogonal rotation $R_C$ and positive channel scale $S_C=\operatorname{Diag}(s_C)$.
For each projection $W$ consuming the latent,
\begin{equation}
    z_C=uR_CS_C,\qquad
    \widetilde W=S_C^{-1}R_C^\top\Gamma W,
    \qquad R_C^\top R_C=I,
    \label{eq:content_transform}
\end{equation}
which exactly preserves the full-precision computation, $z_C\widetilde W=cW$.
The shared latent supplies both keys and values, so the transformed cache is shared by matching and aggregation.
Since weightless RMSNorm is rotation-equivariant, $R_C$ can be fused offline; the learned scale replaces the normalization scale, while $\Gamma$ is absorbed into the compensated consumers (Appendix~\ref{app:content_fusion}).

\paragraph{RoPE path.}
A transformation fused before RoPE must commute with the positional rotations~\citep{roformer}, ruling out unrestricted dense rotations across RoPE frequency pairs.
Appendix~\ref{app:rope_fusion} provides the full fusion derivation and model-specific pairing details.

\begin{proposition}[RoPE-Compatible Equivalent Transformation]
\label{prop:rope_equivalence}
Let $R_P=\operatorname{blockdiag}(R_i)$ with $R_i\in SO(2)$ over RoPE frequency pairs, and let $S_P=\operatorname{blockdiag}(s_iI_2)$ with $s_i>0$.
The reciprocal transformation
\begin{equation}
    Q_P'=Q_PR_PS_P^{-1},\qquad
    K_P'=K_PR_PS_P
    \label{eq:rope_reciprocal_transform}
\end{equation}
preserves the full-precision positional scores,
$Q_P'(K_P')^\top=Q_PK_P^\top$.
Because $R_P$ and $S_P$ commute with the corresponding RoPE rotations, the reciprocal factors can be fused into the pre-RoPE query and key projections without online transformations.
\end{proposition}

\subsection{Function-Aligned Learning Objectives}
\label{subsec:learning_objectives}

Within these offline-fusible spaces, Equation~\ref{eq:finite_path_errors} motivates attention-output reconstruction for the content path and positional QK reconstruction for the RoPE path.

\paragraph{Attention-output reconstruction.}
For the content path, we minimize normalized post-output-projection distortion against the frozen BF16 teacher:
\begin{equation}
    \mathcal L_C
    =\E_{x,\ell}
    \frac{\|Y_\ell(\mathcal Q_C^{R,S}(C),K_P;x)
    -Y_\ell(C,K_P;x)\|_F^2}
    {\max\{\|Y_\ell(C,K_P;x)\|_F^2,\epsilon_C\}}.
    \label{eq:content_objective}
\end{equation}
Here, $\mathcal Q_C^{R,S}$ denotes cache quantization in the learned coordinates with compensated consumers, while the RoPE path remains exact.
Because content errors affect both matching and aggregation, attention-output reconstruction captures their coupled effects and the interaction term that score-only reconstruction would miss.

\paragraph{Positional QK reconstruction.}
RoPE quantization perturbs only the positional scores $Z_{P,h}=Q_{P,h}K_P^\top$, while the content path and value representations remain fixed.
This structure yields a finite bound on output distortion directly in terms of positional-score error.

\begin{proposition}[Finite Output Bound for Positional-QK Perturbation]
\label{prop:qk_output_bound}
Fix the content logits, queries, value representations, and valid causal key sets, with at least one valid key per query.
For an unscaled positional-score perturbation $\Delta Z_{P,h}$ to $Z_{P,h}=Q_{P,h}K_P^\top$, let $\bar V_h=CB_h$ and let $\Omega$ denote the binary valid-pair mask.
Then
\begin{equation}
    \|\Delta Y_P\|_F^2
    \leq
    \frac{\tau^2}{4}
    \left(\sum_h\|\bar V_h\|_2^2\right)
    \sum_h\|\Omega\odot\Delta Z_{P,h}\|_F^2.
    \label{eq:qk_output_bound}
\end{equation}
\end{proposition}

Appendix~\ref{app:qk_bound} provides the proof.
The bound motivates normalized positional QK reconstruction:
\begin{equation}
    \mathcal L_P
    =\E_{x,\ell}
    \frac{\|\Omega\odot(\widetilde Q_{P,\ell}\widehat K_{P,\ell}^{\top}
    -Q_{P,\ell}K_{P,\ell}^{\top})\|_F^2}
    {\max\{\|\Omega\odot Q_{P,\ell}K_{P,\ell}^{\top}\|_F^2,\epsilon_P\}},
    \label{eq:rope_qk_objective}
\end{equation}
where $\widetilde Q_P$ is the compensated, unquantized query and $\widehat K_P$ the transformed quantized key.
Using the actual queries preserves cross-frequency cancellation that key- or frequency-wise reconstruction would miss.
Because equivalent query--key compensation preserves the full-precision scores while leaving the value representations unchanged, minimizing positional QK error directly controls the score-dependent term in the bound.
Appendix~\ref{app:calibration} specifies the quantizer, calibration settings, and optimization procedure.

\subsection{Native Low-Bit MLA Execution}
\label{subsec:kernel_design}

We implement a native low-bit MLA backend on FlashMLA's decode path~\citep{flashmla}, consuming content and RoPE caches in offline-fused coordinates. Unpacking and dequantization are fused into attention without expanding per-head keys and values or materializing a full-history BF16 cache (Figure~\ref{fig:kernel_dataflow}). In the optimized decode kernel, packed data and quantization metadata are buffered separately from reconstructed BF16 tiles, allowing prefetch to overlap with attention computation. Separate metadata storage avoids shared-memory aliasing with reconstructed RoPE tiles, allowing earlier scheduling of RoPE QK computation. Protected BF16 content loads overlap with dequantization of unprotected rows, with synchronization before consumption. Reconstructed content tiles are reused across QK and latent aggregation, and their buffers are overwritten only after all prior consumers complete.
When both the content cache and RoPE key cache use INT4 for low-bit history, a fused writer maintains packed INT4 copies of all tokens and pre-quantization BF16 values for the first four sink tokens and the most recent 128 tokens in both caches. Attention uses exactly one representation per token: BF16 for the union of the protected regions and dequantized INT4 otherwise. Non-sink tokens leaving the recent window switch to their existing packed copies without additional quantization. All valid tokens participate in a single global softmax, preserving full-history attention. Appendix~\ref{app:kernel} details the buffer layouts, pipeline dependencies, and cache lifecycle.

\section{Experimental Evaluation}
\label{sec:experiments}

\begin{table}[t]
\centering
\small
\vspace{-7mm}
\caption{\textbf{General-purpose and information-extraction performance.} Scores (\%, $\uparrow$). CS averages the five commonsense benchmarks; Avg. assigns equal weight to all ten tasks. Bold marks the best quantized score within each precision, including ties. $\dagger$: our MLA adaptations.}
\label{tab:general_results}
\setlength{\tabcolsep}{3.5pt}
\renewcommand{\arraystretch}{1.05}
\setlength{\cmidrulewidth}{0.3pt}
\setlength{\aboverulesep}{1.5pt}
\setlength{\belowrulesep}{1.5pt}
\begin{tabular*}{\linewidth}{@{\extracolsep{\fill}}cl*{7}{r}@{}}
\toprule
\multirow[c]{2}{*}{Precision} & \multirow[c]{2}{*}{Method} & \multicolumn{3}{c}{General-purpose} & \multicolumn{3}{c}{Information extraction} & \multirow[c]{2}{*}{Avg.} \\
\cmidrule(lr){3-5}
\cmidrule(lr){6-8}
 & & CS & MMLU & GSM8K & FDA & SWDE & SQuAD & \\
\midrule
\multicolumn{9}{@{}l}{\textbf{DeepSeek-V2-Lite}} \\
\addlinespace[2pt]
BF16 & --- & 69.76 & 57.90 & 36.92 & 78.13 & 89.20 & 57.10 & 66.80 \\
\cmidrule(lr){1-9}
\multirow{4}{*}{C4R4} & RTN & 67.17 & 51.90 & 16.98 & 65.06 & 84.88 & 55.50 & 61.02 \\
 & SmoothQuant$^{\dagger}$ & 68.23 & 54.02 & 24.56 & 70.69 & 87.85 & 56.07 & 63.44 \\
 & QuaRot$^{\dagger}$ & 69.01 & 55.66 & 30.71 & 72.14 & \textbf{88.12} & 55.06 & 64.67 \\
 & \textbf{QuantMLA} & \textbf{69.92} & \textbf{57.95} & \textbf{35.25} & \textbf{77.77} & \textbf{88.12} & \textbf{56.70} & \textbf{66.54} \\
\cmidrule(lr){1-9}
\multirow{4}{*}{C2R4} & RTN & 62.21 & 40.91 & 3.71 & 30.40 & 70.75 & 44.74 & 50.16 \\
 & SmoothQuant$^{\dagger}$ & 65.93 & 49.40 & 15.31 & 45.10 & 80.20 & 52.51 & 57.22 \\
 & QuaRot$^{\dagger}$ & 64.32 & 45.92 & 12.13 & 40.11 & 75.52 & 46.08 & 54.14 \\
 & \textbf{QuantMLA} & \textbf{69.85} & \textbf{57.29} & \textbf{37.91} & \textbf{60.44} & \textbf{84.52} & \textbf{54.79} & \textbf{64.42} \\
\midrule
\addlinespace[2pt]
\multicolumn{9}{@{}l}{\textbf{Moonlight-16B-A3B}} \\
\addlinespace[2pt]
BF16 & --- & 74.32 & 69.97 & 74.53 & 78.40 & 90.10 & 51.07 & 73.57 \\
\cmidrule(lr){1-9}
\multirow{4}{*}{C4R4} & RTN & 73.94 & 67.99 & 69.60 & 75.59 & \textbf{90.64} & \textbf{51.11} & 72.46 \\
 & SmoothQuant$^{\dagger}$ & 73.39 & 68.45 & 68.99 & 77.22 & 89.29 & 49.87 & 72.08 \\
 & QuaRot$^{\dagger}$ & 74.11 & 67.73 & 71.87 & 74.41 & 90.28 & 50.20 & 72.50 \\
 & \textbf{QuantMLA} & \textbf{74.24} & \textbf{69.71} & \textbf{74.53} & \textbf{78.77} & 89.65 & 50.54 & \textbf{73.44} \\
\cmidrule(lr){1-9}
\multirow{4}{*}{C2R4} & RTN & 69.78 & 60.72 & 41.77 & 47.01 & 82.81 & 36.03 & 61.72 \\
 & SmoothQuant$^{\dagger}$ & 64.95 & 58.99 & 33.81 & 61.52 & 83.35 & \textbf{56.23} & 61.86 \\
 & QuaRot$^{\dagger}$ & 62.72 & 56.16 & 27.37 & 25.50 & 72.73 & 35.36 & 53.07 \\
 & \textbf{QuantMLA} & \textbf{74.22} & \textbf{69.29} & \textbf{75.21} & \textbf{72.50} & \textbf{88.21} & 50.64 & \textbf{72.70} \\
\bottomrule
\end{tabular*}
\vspace{-3mm}
\end{table}

\subsection{Experimental Setup}
\label{subsec:experimental_setup}

\paragraph{Models and tasks.}
We evaluate four MLA families across two capability-matched benchmark suites.
DeepSeek-V2-Lite~\citep{deepseekv2} and Moonlight-16B-A3B~\citep{moonlight} are evaluated on five commonsense benchmarks~\citep{hellaswag,piqa,arc,winogrande}, MMLU~\citep{mmlu}, GSM8K~\citep{gsm8k}, and the FDA, SWDE, and SQuAD recall tasks~\citep{based}.
LongCat-Flash-Lite~\citep{longcat} and GLM-4.7-Flash~\citep{glm47flash} are evaluated on a reasoning-and-code suite~\citep{quantized_reasoning_models} comprising GPQA-Diamond~\citep{gpqa}, MMLU, GSM8K, MATH500~\citep{math}, AIME25, HumanEval~\citep{humaneval}, and LiveCodeBench~\citep{livecodebench}.
We additionally evaluate long-context retrieval on DeepSeek-V2-Lite using RULER S-NIAH-1/2/3~\citep{ruler}, with 500 examples per variant at four context lengths from 4K to 32K tokens.

\paragraph{Baselines and precision.}
To our knowledge, no prior work has reported joint INT4 quantization of MLA's content and RoPE caches, leaving no directly comparable prior method at this precision.
We therefore compare against round-to-nearest quantization (RTN) and two representative transformation-based MLA adaptations, SmoothQuant$^\dagger$~\citep{smoothquant} and QuaRot$^\dagger$~\citep{quarot}, which retain channel scaling and fixed Hadamard rotations, respectively.
Neither original method targets MLA's shared-latent and decoupled-RoPE cache structure; our adaptations operate directly on MLA's native cache representations, with implementation details provided in Appendix~\ref{app:baseline_scope}.
All quantized methods use the same per-token asymmetric affine quantizer with group size 64.
C$x$R$y$ denotes $x$-bit content-cache and $y$-bit RoPE-key-cache precision for low-bit history; C4R4 is the primary setting, while C2R4 further compresses the content cache to INT2.

\paragraph{Calibration.}
With pretrained parameters frozen, we optimize each path-specific transformation for 300 steps on 128 WikiText-2~\citep{wikitext} sequences and select the best checkpoint on 32 disjoint held-out sequences according to the corresponding reconstruction objective.
The objectives are attention-output reconstruction for the content path and positional QK reconstruction for the RoPE path; both learned transformations are fused entirely offline before evaluation.

\subsection{Main Accuracy Results}
\label{subsec:main_results}

\paragraph{General-purpose and extraction tasks.}
Table~\ref{tab:general_results} reports the ten-task suite, with the five commonsense benchmarks detailed in Appendix~\ref{app:full_results}.
At C4R4, QuantMLA remains within $0.26$ and $0.13$ percentage points of BF16 on DeepSeek-V2-Lite and Moonlight-16B-A3B, respectively.
On GSM8K, DeepSeek-V2-Lite reaches $35.25\%$, compared with $16.98\%$, $24.56\%$, and $30.71\%$ for RTN, SmoothQuant$^\dagger$, and QuaRot$^\dagger$, while Moonlight-16B-A3B matches BF16 at $74.53\%$.
Even at C2R4, QuantMLA retains suite averages of $64.42\%$ and $72.70\%$, outperforming the strongest same-precision baselines by $7.20$ and $10.84$ points.

\begin{table}[t]
\centering
\small
\vspace{-5mm}
\caption{\textbf{Reasoning and code performance.} Scores (\%, $\uparrow$). Avg. assigns equal weight to all seven benchmarks. Bold marks the best quantized score within each precision. $\dagger$: our MLA adaptations.}
\label{tab:reasoning_results}
\setlength{\tabcolsep}{3.5pt}
\renewcommand{\arraystretch}{1.05}
\setlength{\cmidrulewidth}{0.3pt}
\setlength{\aboverulesep}{1.5pt}
\setlength{\belowrulesep}{1.5pt}
\begin{tabular*}{\linewidth}{@{\extracolsep{\fill}}cl*{8}{r}@{}}
\toprule
\multirow[c]{3}{*}{Precision} & \multirow[c]{3}{*}{Method} & \multicolumn{5}{c}{Reasoning} & \multicolumn{2}{c}{Code} & \multirow[c]{3}{*}{Avg.} \\
\cmidrule(lr){3-7}
\cmidrule(lr){8-9}
 & & GPQA- & \multirow[c]{2}{*}{MMLU} & \multirow[c]{2}{*}{GSM8K} & MATH & \multirow[c]{2}{*}{AIME25} & Human & LiveCode & \\
 & & Diamond & & & 500 & & Eval & Bench & \\
\midrule
\multicolumn{10}{@{}l}{\textbf{LongCat-Flash-Lite}} \\
\addlinespace[2pt]
BF16 & --- & 46.46 & 80.77 & 71.72 & 57.40 & 70.00 & 73.78 & 49.67 & 64.26 \\
\cmidrule(lr){1-10}
\multirow{4}{*}{C4R4} & RTN & 42.42 & 76.38 & 68.01 & 48.00 & 40.00 & 39.02 & 32.70 & 49.50 \\
 & SmoothQuant$^{\dagger}$ & 42.42 & 78.49 & \textbf{69.75} & 53.20 & 53.33 & 60.37 & 42.75 & 57.19 \\
 & QuaRot$^{\dagger}$ & 39.90 & 78.52 & 64.67 & 54.40 & 56.67 & 56.71 & 42.65 & 56.22 \\
 & \textbf{QuantMLA} & \textbf{44.95} & \textbf{80.51} & \textbf{69.75} & \textbf{54.80} & \textbf{63.33} & \textbf{71.34} & \textbf{49.38} & \textbf{62.01} \\
\cmidrule(lr){1-10}
\multirow{4}{*}{C2R4} & RTN & 29.29 & 70.72 & 50.95 & 39.20 & 26.67 & 27.44 & 24.27 & 38.36 \\
 & SmoothQuant$^{\dagger}$ & 33.84 & 74.55 & 57.62 & 47.00 & 33.33 & 39.63 & 29.57 & 45.08 \\
 & QuaRot$^{\dagger}$ & 28.28 & 64.27 & 30.02 & 24.80 & 6.67 & 35.98 & 16.78 & 29.54 \\
 & \textbf{QuantMLA} & \textbf{44.95} & \textbf{80.57} & \textbf{68.84} & \textbf{56.80} & \textbf{60.00} & \textbf{71.34} & \textbf{49.00} & \textbf{61.64} \\
\midrule
\addlinespace[2pt]
\multicolumn{10}{@{}l}{\textbf{GLM-4.7-Flash}} \\
\addlinespace[2pt]
BF16 & --- & 42.93 & 72.09 & 83.85 & 56.40 & 63.33 & 36.59 & 32.04 & 55.32 \\
\cmidrule(lr){1-10}
\multirow{4}{*}{C4R4} & RTN & 37.88 & 70.79 & 81.27 & 51.20 & \textbf{56.67} & 29.88 & 29.67 & 51.05 \\
 & SmoothQuant$^{\dagger}$ & 36.87 & 70.60 & 81.12 & 51.00 & 53.33 & 31.71 & 31.47 & 50.87 \\
 & QuaRot$^{\dagger}$ & 34.85 & 70.84 & 82.41 & 52.80 & 46.67 & 28.66 & 30.62 & 49.55 \\
 & \textbf{QuantMLA} & \textbf{38.89} & \textbf{71.95} & \textbf{83.70} & \textbf{57.40} & \textbf{56.67} & \textbf{34.76} & \textbf{32.89} & \textbf{53.75} \\
\cmidrule(lr){1-10}
\multirow{4}{*}{C2R4} & RTN & 30.81 & 64.81 & 71.19 & 42.40 & 36.67 & 31.71 & 21.80 & 42.77 \\
 & SmoothQuant$^{\dagger}$ & 31.82 & 64.48 & 73.62 & 45.40 & \textbf{46.67} & 29.27 & 25.21 & 45.21 \\
 & QuaRot$^{\dagger}$ & 28.79 & 53.64 & 51.10 & 34.20 & 26.67 & 6.10 & 18.10 & 31.23 \\
 & \textbf{QuantMLA} & \textbf{39.39} & \textbf{71.50} & \textbf{82.94} & \textbf{55.00} & \textbf{46.67} & \textbf{34.15} & \textbf{30.33} & \textbf{51.43} \\
\bottomrule
\end{tabular*}
\end{table}

\begin{figure}[t]
    \centering
    \vspace{-3mm}
    \includegraphics[width=\textwidth]{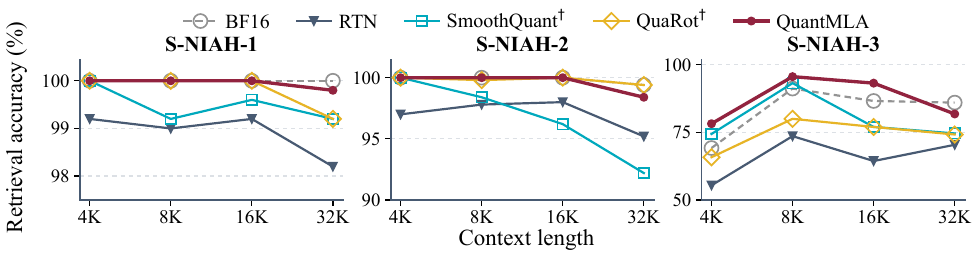}
    \vspace{-8mm}
    \caption{\textbf{Retrieval as the cached history grows.}
    DeepSeek-V2-Lite on RULER S-NIAH-1/2/3 at C4R4, with BF16 as reference.
    Each point evaluates 500 examples; method markers follow Table~\ref{tab:general_results}.}
    \label{fig:niah_results}
    \vspace{-3mm}
\end{figure}

\paragraph{Reasoning and code tasks.}
Table~\ref{tab:reasoning_results} shows a similar trend on the reasoning-and-code suite.
At C4R4, QuantMLA achieves averages of $62.01\%$ and $53.75\%$ on LongCat-Flash-Lite and GLM-4.7-Flash, exceeding the strongest same-precision baselines by $4.82$ and $2.70$ points, respectively.
Several tasks remain close to BF16, including LiveCodeBench on LongCat-Flash-Lite ($0.29$ points lower) and GSM8K on GLM-4.7-Flash ($0.15$ points lower).
Under C2R4, LongCat-Flash-Lite still retains $71.34\%$ on HumanEval and $49.00\%$ on LiveCodeBench.
GLM-4.7-Flash reaches a $51.43\%$ average, $6.22$ points above the strongest same-precision baseline, while the larger task-to-task variation indicates greater sensitivity to extreme content-cache compression on some benchmarks.

\paragraph{Long-context retrieval.}
Figure~\ref{fig:niah_results} evaluates C4R4 retrieval from 4K to 32K tokens.
QuantMLA remains close to BF16 on S-NIAH-1 and S-NIAH-2, averaging $99.95\%$ and $99.60\%$, respectively.
On the more challenging S-NIAH-3, it averages $87.20\%$, exceeding the strongest quantized baseline by $7.40$ percentage points and outperforming all quantized baselines at every tested length.

\vspace{-2mm}
\subsection{Ablation Studies}
\label{subsec:ablations}
\vspace{-8mm}

\noindent
\begin{minipage}[t]{0.48\textwidth}
\vspace{\abovecaptionskip}
Table~\ref{tab:ablations} evaluates QuantMLA's component contributions and compares its unprotected C4R4 representation with representative baselines on the full DeepSeek-V2-Lite MMLU test set.
Starting from RTN, the content- and RoPE-path transformations improve accuracy by $1.22$ and $3.19$ percentage points, respectively, raising the unprotected configuration from $51.90\%$ to $56.31\%$ before any BF16 protection.
This transformation-only configuration outperforms SmoothQuant$^\dagger$ and QuaRot$^\dagger$ by $2.29$ and $0.65$ points, respectively.
Building on this strong low-bit representation, the efficiently supported mixed-precision cache policy adds a further $1.64$ points, bringing the complete QuantMLA configuration to $57.95\%$, comparable to BF16 at $57.90\%$.
\end{minipage}\hfill
\begin{minipage}[t]{0.49\textwidth}
\vspace{0pt}
\centering
\captionof{table}{\textbf{Baseline comparison and component ablation at C4R4.}
Scores on DeepSeek-V2-Lite MMLU; parenthesized values denote incremental gains within the QuantMLA sequence.}
\label{tab:ablations}
\small
\begin{tabular*}{\linewidth}{@{\extracolsep{\fill}}lr@{}}
\toprule
Configuration & MMLU (\%) $\uparrow$ \\
\midrule
BF16 & 57.90 \\
\midrule
\multicolumn{2}{l}{\textit{Unprotected C4R4 baselines}} \\
RTN & 51.90 \\
SmoothQuant$^\dagger$ & 54.02 \\
QuaRot$^\dagger$ & 55.66 \\
\midrule
\multicolumn{2}{l}{\textit{QuantMLA cumulative components}} \\
RTN & 51.90 \\
+ Content-path transformation & 53.12 ($+1.22$) \\
+ RoPE-path transformation & 56.31 ($+3.19$) \\
+ Mixed-precision cache policy & \textbf{57.95} ($+1.64$) \\
\bottomrule
\end{tabular*}
\end{minipage}
\par

\vspace{-2mm}
\subsection{System Efficiency}
\label{subsec:efficiency}

\begin{figure}[t]
    \centering
    \vspace{-8mm}
    \includegraphics[width=\textwidth]{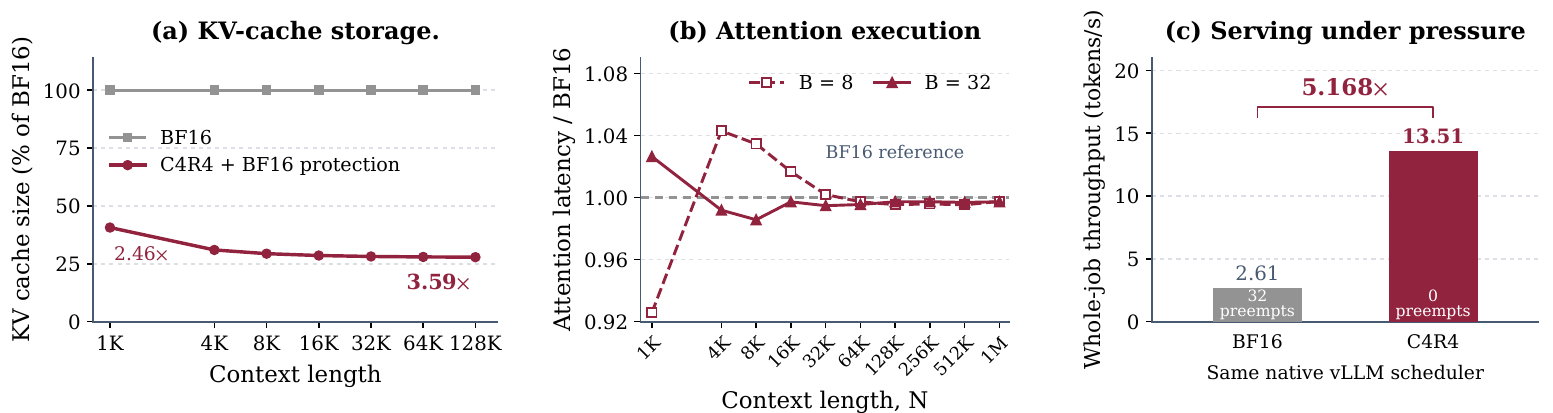}
    \vspace{-9mm}
    \caption{\textbf{KV-cache storage and inference efficiency of C4R4 with sink4/local128.} (a) Persistent KV-cache allocation, including BF16 buffers and metadata. (b) Attention latency at batch sizes 8 and 32, normalized to paired measurements with an unquantized BF16 cache. (c) Whole-job output throughput on eight GPUs under the same native vLLM scheduling policy; labels report preemption counts. Panel (c) compares one C4R4 run against a historical BF16 baseline.}
    \vspace{-3mm}
    \label{fig:efficiency}
\end{figure}

\vspace{-3mm}
\paragraph{KV-cache storage.}
With the mixed-precision cache policy, C4R4 achieves $3.22\times$ and $3.59\times$ persistent KV-cache compression at 4K and 128K, respectively (Figure~\ref{fig:efficiency}(a)). The accounting includes BF16 buffers, redundant packed copies, and page mappings (Appendix~\ref{app:kernel}).

\vspace{-3mm}
\paragraph{Native attention execution.}
Across tested lengths from 1K to 1M at batch sizes 8 and 32, C4R4 achieves near-BF16 attention latency, ranging from $7.42\%$ lower to $4.29\%$ higher, with differences below $0.5\%$ from 64K onward (Figure~\ref{fig:efficiency}(b); Appendix~\ref{app:system_efficiency}). With L2 flushed before each graph sample, CUDA Graph timing spans attention-main through split-result combination, including unpacking and dequantization but excluding cache writing and metadata construction.

\vspace{-3mm}
\paragraph{Serving under memory pressure.}
For five simultaneous requests with 128K input and 1K output tokens on eight GPUs, BF16 incurs 32 preemption events and C4R4 none under the same native vLLM scheduling policy~\citep{vllm}. Whole-job output throughput increases from $2.61$ to $13.51$ tokens/s ($5.168\times$; Figure~\ref{fig:efficiency}(c)), including prefill and recomputation. This gain is specific to the evaluated cache-pressure workload (Appendix~\ref{app:serving}).

\vspace{-2mm}
\section{Conclusion}
\label{sec:conclusion}
\vspace{-3mm}
QuantMLA connects MLA's dual-path asymmetry to low-bit cache learning.
Its error analysis motivates path-specific offline-fusible transformations and function-aligned objectives by explaining unequal output distortion under matched cache error.
Across four MLA families, joint INT4 caching remains close to BF16, while INT2 content caching extends the accuracy--memory trade-off.
Offline fusion with native low-bit execution achieves $3.59\times$ cache compression at 128K and $5.168\times$ BF16 whole-job throughput under cache pressure.

\clearpage

\subsection*{AI use statement}
Generative AI tools were used solely for language editing, including grammar, spelling, word choice, and minor improvements to clarity and consistency. All technical content, experimental results, analyses, and conclusions were developed and verified by the authors. The authors take full responsibility for the final manuscript.

\subsection*{Ethics statement}
Low-bit caching can alter model behavior even when aggregate benchmark accuracy is preserved.
Deployments should therefore reassess safety, calibration, and failure modes under the exact model, precision, and serving configuration.
Our study uses public models and standard public benchmarks and involves no human subjects or collection of personal data.

\subsection*{Reproducibility statement}
Appendices~C and~D specify the quantization operators, calibration protocol, hyperparameters, baseline adaptations, and evaluation settings.
Appendix~E provides the detailed commonsense results underlying the aggregated main-table scores, while Appendix~B.3 reports controlled validation of the proposed error model.
The derivations supporting the error analysis and the algebraic fusion of the learned transformations appear in Appendices~B.2 and~C, respectively.

\bibliography{references}

@inproceedings{quantized_reasoning_models,
  title = {Quantization Hurts Reasoning? An Empirical Study on Quantized Reasoning Models},
  author = {Liu, Ruikang and Sun, Yuxuan and Zhang, Manyi and Bai, Haoli and Yu, Xianzhi and Yu, Tiezheng and Yuan, Chun and Hou, Lu},
  booktitle = {Conference on Language Modeling},
  year = {2025},
  url = {https://arxiv.org/abs/2504.04823}
}

@inproceedings{linear_attention,
  title = {Transformers are {RNN}s: Fast Autoregressive Transformers with Linear Attention},
  author = {Katharopoulos, Angelos and Vyas, Apoorv and Pappas, Nikolaos and Fleuret, Fran{\c{c}}ois},
  booktitle = {Proceedings of the 37th International Conference on Machine Learning},
  pages = {5156--5165},
  year = {2020},
  url = {https://proceedings.mlr.press/v119/katharopoulos20a.html}
}

@article{hla,
  title = {A Systematic Analysis of Hybrid Linear Attention},
  author = {Wang, Dustin and Zhu, Rui-Jie and Abreu, Steven and Shan, Yong and Kergan, Taylor and Pan, Yuqi and Chou, Yuhong and Li, Zheng and Wu, Jibin and Zhang, Ge and Huang, Wenhao and Eshraghian, Jason},
  journal = {arXiv preprint arXiv:2507.06457},
  year = {2025},
  note = {Version 2, June 2026},
  url = {https://arxiv.org/abs/2507.06457}
}

@misc{kimi_k3,
  title = {{Kimi K3}: Open Frontier Intelligence},
  author = {{Moonshot AI}},
  year = {2026},
  howpublished = {Official model release and technical report},
  url = {https://github.com/MoonshotAI/Kimi-K3}
}

@misc{longcat2,
  title = {{LongCat-2.0}},
  author = {{Meituan LongCat Team}},
  year = {2026},
  howpublished = {Official model release},
  url = {https://github.com/meituan-longcat/LongCat-2.0}
}

@misc{glm53,
  title = {{GLM-5.3}},
  author = {{Z.ai}},
  year = {2026},
  howpublished = {Official model card},
  url = {https://huggingface.co/zai-org/GLM-5.3}
}

@inproceedings{vllm,
  title = {Efficient Memory Management for Large Language Model Serving with {PagedAttention}},
  author = {Kwon, Woosuk and Li, Zhuohan and Zhuang, Siyuan and Sheng, Ying and Zheng, Lianmin and Yu, Cody Hao and Gonzalez, Joseph E. and Zhang, Hao and Stoica, Ion},
  booktitle = {Proceedings of the 29th Symposium on Operating Systems Principles},
  year = {2023},
  url = {https://arxiv.org/abs/2309.06180}
}

@article{deepseekv2,
  title        = {DeepSeek-V2: A Strong, Economical, and Efficient Mixture-of-Experts Language Model},
  author       = {{DeepSeek-AI}},
  journal      = {arXiv preprint arXiv:2405.04434},
  year         = {2024}
}

@inproceedings{kivi,
  title        = {{KIVI}: A Tuning-Free Asymmetric 2bit Quantization for {KV} Cache},
  author       = {Liu, Zirui and Yuan, Jiayi and Jin, Hongye and Zhong, Shaochen and Xu, Zhaozhuo and Braverman, Vladimir and Chen, Beidi and Hu, Xia},
  booktitle    = {International Conference on Machine Learning},
  year         = {2024}
}

@inproceedings{kvquant,
  title        = {{KVQuant}: Towards 10 Million Context Length {LLM} Inference with {KV} Cache Quantization},
  author       = {Hooper, Coleman and Kim, Sehoon and Mohammadzadeh, Hiva and Mahoney, Michael W. and Shao, Yakun Sophia and Keutzer, Kurt and Gholami, Amir},
  booktitle    = {Advances in Neural Information Processing Systems},
  year         = {2024}
}

@inproceedings{smoothquant,
  title        = {{SmoothQuant}: Accurate and Efficient Post-Training Quantization for Large Language Models},
  author       = {Xiao, Guangxuan and Lin, Ji and Seznec, Mickael and Wu, Hao and Demouth, Julien and Han, Song},
  booktitle    = {International Conference on Machine Learning},
  year         = {2023}
}

@inproceedings{omniquant,
  title        = {{OmniQuant}: Omnidirectionally Calibrated Quantization for Large Language Models},
  author       = {Shao, Wenqi and Chen, Mengzhao and Zhang, Zhaoyang and Xu, Peng and Zhao, Lirui and Li, Zhiqian and Zhang, Kaipeng and Gao, Peng and Qiao, Yu and Luo, Ping},
  booktitle    = {International Conference on Learning Representations},
  year         = {2024}
}

@inproceedings{quarot,
  title        = {{QuaRot}: Outlier-Free 4-Bit Inference in Rotated {LLM}s},
  author       = {Ashkboos, Saleh and Mohtashami, Amirkeivan and Croci, Maximilian L. and Li, Bo and Cameron, Pashmina and Jaggi, Martin and Alistarh, Dan and Hoefler, Torsten and Hensman, James},
  booktitle    = {Advances in Neural Information Processing Systems},
  year         = {2024}
}

@inproceedings{spinquant,
  title        = {{SpinQuant}: {LLM} Quantization with Learned Rotations},
  author       = {Liu, Zechun and Zhao, Changsheng and Fedorov, Igor and Soran, Bilge and Choudhary, Dhruv and Krishnamoorthi, Raghuraman and Chandra, Vikas and Tian, Yuandong and Blankevoort, Tijmen},
  booktitle    = {International Conference on Learning Representations},
  year         = {2025}
}

@inproceedings{flatquant,
  title = {{FlatQuant}: Flatness Matters for {LLM} Quantization},
  author = {Sun, Yuxuan and Liu, Ruikang and Bai, Haoli and Bao, Han and Zhao, Kang and Li, Yuening and Hu, Jiaxin and Yu, Xianzhi and Hou, Lu and Yuan, Chun and Jiang, Xin and Liu, Wulong and Yao, Jun},
  booktitle = {International Conference on Machine Learning},
  year = {2025},
  url = {https://arxiv.org/abs/2410.09426}
}

@misc{flashmla,
  title = {{FlashMLA}: Efficient Multi-head Latent Attention Kernels},
  author = {{DeepSeek-AI}},
  year = {2025},
  url = {https://github.com/deepseek-ai/FlashMLA},
  note = {Accessed September 22, 2026; layouts are model-version specific}
}

@inproceedings{rotatekv,
  title        = {{RotateKV}: Accurate and Robust 2-Bit {KV} Cache Quantization for {LLM}s via Outlier-Aware Adaptive Rotations},
  author       = {Su, Zunhai and Wei, Hanyu and Chen, Zhe and Shen, Wang and Li, Linge and Yu, Huangqi and Yuan, Kehong},
  booktitle    = {Proceedings of the Thirty-Fourth International Joint Conference on Artificial Intelligence},
  pages        = {6200--6208},
  doi          = {10.24963/ijcai.2025/690},
  url          = {https://www.ijcai.org/proceedings/2025/690},
  year         = {2025}
}

@article{oscar,
  title        = {{OScaR}: The Occam's Razor for Extreme {KV} Cache Quantization in {LLM}s and Beyond},
  author       = {Su, Zunhai and Yang, Rui and Zhang, Chao and Liu, Yaxiu and Zhang, Yifan and Wu, Wei and Xiong, Jing and Du, Dayou and Zhuang, Xialie and Qian, Yulei and Xie, Yuchen and Wu, Yik-Chung and Yang, Hongxia and Wong, Ngai},
  journal      = {arXiv preprint arXiv:2605.19660},
  year         = {2026}
}

@article{snapmla,
  title        = {{SnapMLA}: Efficient Long-Context {MLA} Decoding via Hardware-Aware {FP8} Quantized Pipelining},
  author       = {Zhang, Yifan and Su, Zunhai and Hu, Shuhao and Yang, Rui and Wu, Wei and Qian, Yulei and Xie, Yuchen and Cai, Xunliang},
  journal      = {arXiv preprint arXiv:2602.10718},
  year         = {2026}
}

@article{moonlight,
  title        = {Muon is Scalable for {LLM} Training},
  author       = {Liu, Jingyuan and Su, Jianlin and Yao, Xingcheng and Jiang, Zhejun and Lai, Guokun and Du, Yulun and Qin, Yidao and Xu, Weixin and Lu, Enzhe and Yan, Junjie and Chen, Yanru and Zheng, Huabin and Liu, Yibo and Liu, Shaowei and Yin, Bohong and He, Weiran and Zhu, Han and Wang, Yuzhi and Wang, Jianzhou and Dong, Mengnan and Zhang, Zheng and Kang, Yongsheng and Zhang, Hao and Xu, Xinran and Zhang, Yutao and Wu, Yuxin and Zhou, Xinyu and Yang, Zhilin},
  journal      = {arXiv preprint arXiv:2502.16982},
  year         = {2025}
}

@misc{longcat,
  title        = {{LongCat-Flash-Lite}},
  author       = {{Meituan LongCat Team}},
  howpublished = {Model card},
  url          = {https://huggingface.co/meituan-longcat/LongCat-Flash-Lite},
  year         = {2026}
}

@misc{glm47flash,
  title        = {{GLM-4.7-Flash}},
  author       = {{Z.ai}},
  year         = {2026},
  howpublished = {Model release},
  url          = {https://huggingface.co/zai-org/GLM-4.7-Flash}
}

@inproceedings{rmsnorm,
  title        = {Root Mean Square Layer Normalization},
  author       = {Zhang, Biao and Sennrich, Rico},
  booktitle    = {Advances in Neural Information Processing Systems},
  year         = {2019}
}

@article{roformer,
  title        = {{RoFormer}: Enhanced Transformer with Rotary Position Embedding},
  author       = {Su, Jianlin and Lu, Yu and Pan, Shengfeng and Murtadha, Ahmed and Wen, Bo and Liu, Yunfeng},
  journal      = {Neurocomputing},
  volume       = {568},
  pages        = {127063},
  year         = {2024}
}

@inproceedings{wikitext,
  title        = {Pointer Sentinel Mixture Models},
  author       = {Merity, Stephen and Xiong, Caiming and Bradbury, James and Socher, Richard},
  booktitle    = {International Conference on Learning Representations},
  year         = {2017}
}

@article{t5,
  title        = {Exploring the Limits of Transfer Learning with a Unified Text-to-Text Transformer},
  author       = {Raffel, Colin and Shazeer, Noam and Roberts, Adam and Lee, Katherine and Narang, Sharan and Matena, Michael and Zhou, Yanqi and Li, Wei and Liu, Peter J.},
  journal      = {Journal of Machine Learning Research},
  volume       = {21},
  number       = {140},
  pages        = {1--67},
  year         = {2020}
}

@inproceedings{mmlu,
  title        = {Measuring Massive Multitask Language Understanding},
  author       = {Hendrycks, Dan and Burns, Collin and Basart, Steven and Zou, Andy and Mazeika, Mantas and Song, Dawn and Steinhardt, Jacob},
  booktitle    = {International Conference on Learning Representations},
  year         = {2021}
}

@article{gsm8k,
  title        = {Training Verifiers to Solve Math Word Problems},
  author       = {Cobbe, Karl and Kosaraju, Vineet and Bavarian, Mohammad and Chen, Mark and Jun, Heewoo and Kaiser, Lukasz and Plappert, Matthias and Tworek, Jerry and Hilton, Jacob and Nakano, Reiichiro and Hesse, Christopher and Schulman, John},
  journal      = {arXiv preprint arXiv:2110.14168},
  year         = {2021},
  url          = {https://arxiv.org/abs/2110.14168}
}

@article{humaneval,
  title        = {Evaluating Large Language Models Trained on Code},
  author       = {Chen, Mark and others},
  journal      = {arXiv preprint arXiv:2107.03374},
  year         = {2021}
}

@inproceedings{math,
  title        = {Measuring Mathematical Problem Solving With the {MATH} Dataset},
  author       = {Hendrycks, Dan and Burns, Collin and Kadavath, Saurav and Arora, Akul and Basart, Steven and Tang, Eric and Song, Dawn and Steinhardt, Jacob},
  booktitle    = {Proceedings of the Neural Information Processing Systems Track on Datasets and Benchmarks},
  year         = {2021},
  url          = {https://arxiv.org/abs/2103.03874}
}

@inproceedings{ruler,
  title        = {{RULER}: What's the Real Context Size of Your Long-Context Language Models?},
  author       = {Hsieh, Cheng-Ping and Sun, Simeng and Kriman, Samuel and Acharya, Shantanu and Rekesh, Dima and Jia, Fei and Zhang, Yang and Ginsburg, Boris},
  booktitle    = {Conference on Language Modeling},
  year         = {2024}
}

@inproceedings{gpqa,
  title        = {{GPQA}: A Graduate-Level Google-Proof Q\&A Benchmark},
  author       = {Rein, David and others},
  booktitle    = {Conference on Language Modeling},
  year         = {2024}
}

@inproceedings{livecodebench,
  title        = {{LiveCodeBench}: Holistic and Contamination Free Evaluation of Large Language Models for Code},
  author       = {Jain, Naman and Han, King and Gu, Alex and Li, Wen-Ding and Yan, Fanjia and Zhang, Tianjun and Wang, Sida and Solar-Lezama, Armando and Sen, Koushik and Stoica, Ion},
  booktitle    = {International Conference on Learning Representations},
  year         = {2025}
}

@article{deepseekv3,
  title = {{DeepSeek-V3 Technical Report}},
  author = {{DeepSeek-AI} and others},
  journal = {arXiv preprint arXiv:2412.19437},
  year = {2024},
  url = {https://arxiv.org/abs/2412.19437}
}

@article{kimi_k2,
  title = {{Kimi K2: Open Agentic Intelligence}},
  author = {{Kimi Team} and others},
  journal = {arXiv preprint arXiv:2507.20534},
  year = {2025},
  url = {https://arxiv.org/abs/2507.20534}
}

@article{kimi_linear,
  title = {{Kimi Linear: An Expressive, Efficient Attention Architecture}},
  author = {{Kimi Team} and others},
  journal = {arXiv preprint arXiv:2510.26692},
  year = {2025},
  url = {https://arxiv.org/abs/2510.26692}
}

@article{nsa,
  title = {{Native Sparse Attention: Hardware-Aligned and Natively Trainable Sparse Attention}},
  author = {Yuan, Jingyang and others},
  journal = {arXiv preprint arXiv:2502.11089},
  year = {2025},
  url = {https://arxiv.org/abs/2502.11089}
}

@article{rptq,
  title = {{RPTQ: Reorder-based Post-training Quantization for Large Language Models}},
  author = {Yuan, Zhihang and Niu, Lin and Liu, Jiawei and Liu, Wenyu and Wang, Xinggang and Shang, Yuzhang and Sun, Guangyu and Wu, Qiang and Wu, Jiaxiang and Wu, Bingzhe},
  journal = {arXiv preprint arXiv:2304.01089},
  year = {2023},
  url = {https://arxiv.org/abs/2304.01089}
}

@article{longcat_report,
  title = {{LongCat-Flash Technical Report}},
  author = {{Meituan LongCat Team} and others},
  journal = {arXiv preprint arXiv:2509.01322},
  year = {2025},
  url = {https://arxiv.org/abs/2509.01322}
}

@article{hellaswag,
  title = {{HellaSwag: Can a Machine Really Finish Your Sentence?}},
  author = {Zellers, Rowan and Holtzman, Ari and Bisk, Yonatan and Farhadi, Ali and Choi, Yejin},
  journal = {arXiv preprint arXiv:1905.07830},
  year = {2019},
  url = {https://arxiv.org/abs/1905.07830}
}

@article{piqa,
  title = {{PIQA: Reasoning about Physical Commonsense in Natural Language}},
  author = {Bisk, Yonatan and Zellers, Rowan and Le Bras, Ronan and Gao, Jianfeng and Choi, Yejin},
  journal = {arXiv preprint arXiv:1911.11641},
  year = {2019},
  url = {https://arxiv.org/abs/1911.11641}
}

@article{arc,
  title = {{Think you have Solved Question Answering? Try ARC, the AI2 Reasoning Challenge}},
  author = {Clark, Peter and Cowhey, Isaac and Etzioni, Oren and Khot, Tushar and Sabharwal, Ashish and Schoenick, Carissa and Tafjord, Oyvind},
  journal = {arXiv preprint arXiv:1803.05457},
  year = {2018},
  url = {https://arxiv.org/abs/1803.05457}
}

@article{winogrande,
  title = {{WinoGrande: An Adversarial Winograd Schema Challenge at Scale}},
  author = {Sakaguchi, Keisuke and Le Bras, Ronan and Bhagavatula, Chandra and Choi, Yejin},
  journal = {arXiv preprint arXiv:1907.10641},
  year = {2019},
  url = {https://arxiv.org/abs/1907.10641}
}

@article{squad,
  title = {{SQuAD: 100,000+ Questions for Machine Comprehension of Text}},
  author = {Rajpurkar, Pranav and Zhang, Jian and Lopyrev, Konstantin and Liang, Percy},
  journal = {arXiv preprint arXiv:1606.05250},
  year = {2016},
  url = {https://arxiv.org/abs/1606.05250}
}

@article{deepseekr1,
  title = {{DeepSeek-R1: Incentivizing Reasoning Capability in LLMs via Reinforcement Learning}},
  author = {{DeepSeek-AI} and others},
  journal = {arXiv preprint arXiv:2501.12948},
  year = {2025},
  url = {https://arxiv.org/abs/2501.12948}
}

@article{skvq,
  title = {{SKVQ}: Sliding-window Key and Value Cache Quantization for Large Language Models},
  author = {Duanmu, Haojie and Yuan, Zhihang and Li, Xiuhong and Duan, Jiangfei and Zhang, Xingcheng and Lin, Dahua},
  journal = {arXiv preprint arXiv:2405.06219},
  year = {2024},
  url = {https://arxiv.org/abs/2405.06219}
}

@article{duquant,
  title = {{DuQuant}: Distributing Outliers via Dual Transformation Makes Stronger Quantized {LLM}s},
  author = {Lin, Haokun and Xu, Haobo and Wu, Yichen and Cui, Jingzhi and Zhang, Yingtao and Mou, Linzhan and Song, Linqi and Sun, Zhenan and Wei, Ying},
  journal = {arXiv preprint arXiv:2406.01721},
  year = {2024},
  url = {https://arxiv.org/abs/2406.01721}
}

@article{ostquant,
  title = {{OSTQuant}: Refining Large Language Model Quantization with Orthogonal and Scaling Transformations for Better Distribution Fitting},
  author = {Hu, Xing and Cheng, Yuan and Yang, Dawei and Xu, Zukang and Yuan, Zhihang and Yu, Jiangyong and Xu, Chen and Jiang, Zhe and Zhou, Sifan},
  journal = {arXiv preprint arXiv:2501.13987},
  year = {2025},
  url = {https://arxiv.org/abs/2501.13987}
}

@article{fptquant,
  title = {{FPTQuant}: Function-Preserving Transforms for {LLM} Quantization},
  author = {van Breugel, Boris and Bondarenko, Yelysei and Whatmough, Paul and Nagel, Markus},
  journal = {arXiv preprint arXiv:2506.04985},
  year = {2025},
  url = {https://arxiv.org/abs/2506.04985}
}

@article{turboquant,
  title = {{TurboQuant}: Online Vector Quantization with Near-optimal Distortion Rate},
  author = {Zandieh, Amir and Daliri, Majid and Hadian, Majid and Mirrokni, Vahab},
  journal = {arXiv preprint arXiv:2504.19874},
  year = {2025},
  url = {https://arxiv.org/abs/2504.19874}
}

@article{polarquant,
  title = {{PolarQuant}: Quantizing {KV} Caches with Polar Transformation},
  author = {Han, Insu and Kacham, Praneeth and Karbasi, Amin and Mirrokni, Vahab and Zandieh, Amir},
  journal = {arXiv preprint arXiv:2502.02617},
  year = {2025},
  url = {https://arxiv.org/abs/2502.02617}
}

@inproceedings{based,
  title = {Simple linear attention language models balance the recall-throughput tradeoff},
  author = {Arora, Simran and Eyuboglu, Sabri and Zhang, Michael and Timalsina, Aman and Alberti, Silas and Zou, James and Rudra, Atri and Re, Christopher},
  booktitle = {Proceedings of the 41st International Conference on Machine Learning},
  volume = {235},
  pages = {1763--1840},
  year = {2024},
  publisher = {PMLR},
  url = {https://proceedings.mlr.press/v235/arora24a.html}
}

@article{oscar_spectral,
  title = {{OSCAR}: Offline Spectral Covariance-Aware Rotation for 2-bit {KV} Cache Quantization},
  author = {Zhou, Zhongzhu and Zhuang, Donglin and Li, Jisen and Chen, Ziyan and Song, Shuaiwen Leon and Athiwaratkun, Ben and Wu, Xiaoxia},
  journal = {arXiv preprint arXiv:2605.17757},
  year = {2026},
  url = {https://arxiv.org/abs/2605.17757}
}
\bibliographystyle{iclr2027_conference}

\appendix
\clearpage

\section*{Appendix Contents}
\startcontents[appendix]

\begingroup
\titlecontents{section}[0pt]
  {\addvspace{0.3em}\bfseries}
  {\thecontentslabel\quad}
  {}
  {\hfill\contentspage}
\printcontents[appendix]{}{1}{\setcounter{tocdepth}{2}}
\endgroup

\clearpage

\section{Related Work}
\label{app:related_work}

\subsection{Transformation-Based LLM Quantization}
\label{subsec:equivalent_transformations}

Equivalent transformations improve quantization by reshaping representations while preserving full-precision computation.
Early methods redistribute quantization difficulty through scaling or reordering: SmoothQuant redistributes activation outliers through equivalent scaling, while OmniQuant jointly learns equivalent scaling and weight clipping~\citep{smoothquant,omniquant}; RPTQ and SKVQ reorder channels to reduce within-group range variation~\citep{rptq,skvq}.
Orthogonal transforms suppress outliers and redistribute energy, from randomized Hadamard rotations in QuaRot to learned rotations in SpinQuant and rotation--permutation compositions in DuQuant~\citep{quarot,spinquant,duquant}.
More expressive transformation families include affine transforms in FlatQuant, orthogonal--scaling compositions in OSTQuant, and learned pre-RoPE and value transformations in FPTQuant~\citep{flatquant,ostquant,fptquant}.
RotateKV extends rotation-based quantization to KV caches through outlier-aware pre-RoPE transformations, while OScaR combines canalized rotation with token scaling~\citep{rotatekv,oscar}.
These methods are primarily developed for conventional explicit KV representations and do not directly accommodate MLA's latent-cache structure.
In MLA, the shared content latent jointly supplies keys and values, whereas the decoupled RoPE key cache stores positional keys under distinct rotary-compatibility constraints, yielding different offline-fusible transformation spaces and functional roles.
\method{} addresses this architecture-specific asymmetry through path-specific offline-fusible transformation spaces and function-aligned objectives.

\subsection{Low-Bit KV Cache Quantization}
\label{subsec:kv_cache_quantization}

Low-bit KV-cache quantization exploits the heterogeneous statistics of keys and values.
KIVI uses per-channel key and per-token value quantization, while KVQuant introduces pre-RoPE key quantization and outlier-aware coding~\citep{kivi,kvquant}.
Later methods reshape or recode cached representations: RotateKV smooths key outliers through rotation and reordering, while TurboQuant and PolarQuant introduce transformation or coding structures for lower-distortion compression~\citep{rotatekv,turboquant,polarquant}.
Methods targeting more aggressive compression include OSCAR, which derives attention-aware spectral rotations from query--key and value-side covariance, and OScaR, which combines canalized rotation with token scaling to mitigate token-norm imbalance under INT2 quantization~\citep{oscar_spectral,oscar}.
MLA changes the quantization target: a shared content latent jointly supplies content keys and values, while a separate cache stores the decoupled RoPE keys.
FlashMLA's DeepSeek-V3.2 layout adopts an FP8-content/BF16-RoPE split, while SnapMLA co-designs content-cache quantization with attention execution~\citep{flashmla,snapmla}.
These MLA-specific approaches neither model how quantization errors propagate through the two paths nor jointly optimize low-bit representations for both caches.
\method{} closes this gap with dual-path error modeling and function-aligned transformation learning for joint low-bit caching.
\section{Additional Dual-Path Error Analysis}
\label{app:error_analysis}

This appendix complements Section~\ref{sec:functional_risk} with the matched-error protocol and complete six-model results, the derivation of error propagation and amplification, and predictive and interventional validation of the operator--error model.

\subsection{Matched-Error Protocol and Six-Model Results}
\label{app:matched_error}

For each layer and input, we compute the quantization error induced by INT4 affine quantization and rescale it to cache NMSE $0.01$ without changing its direction.
We perturb one path's cache at a time, keeping the other path and model parameters fixed, and measure distortion after the attention output projection.

\paragraph{Inputs and coverage.}
Each model is evaluated on WikiText~\citep{wikitext} and C4~\citep{t5} with three fixed seeds, yielding six dataset--seed cells with eight non-overlapping 512-token sequences per cell.
The study covers 285 attention layers: 27, 61, 61, 61, 28, and 47 for DeepSeek-V2-Lite, DeepSeek-V3-Base, DeepSeek-R1, Kimi-K2-Instruct, LongCat-Flash-Lite, and GLM-4.7-Flash, respectively~\citep{deepseekv2,deepseekv3,deepseekr1,kimi_k2,longcat,glm47flash}.

\begin{figure}[!htbp]
    \centering
    \includegraphics[width=\textwidth]{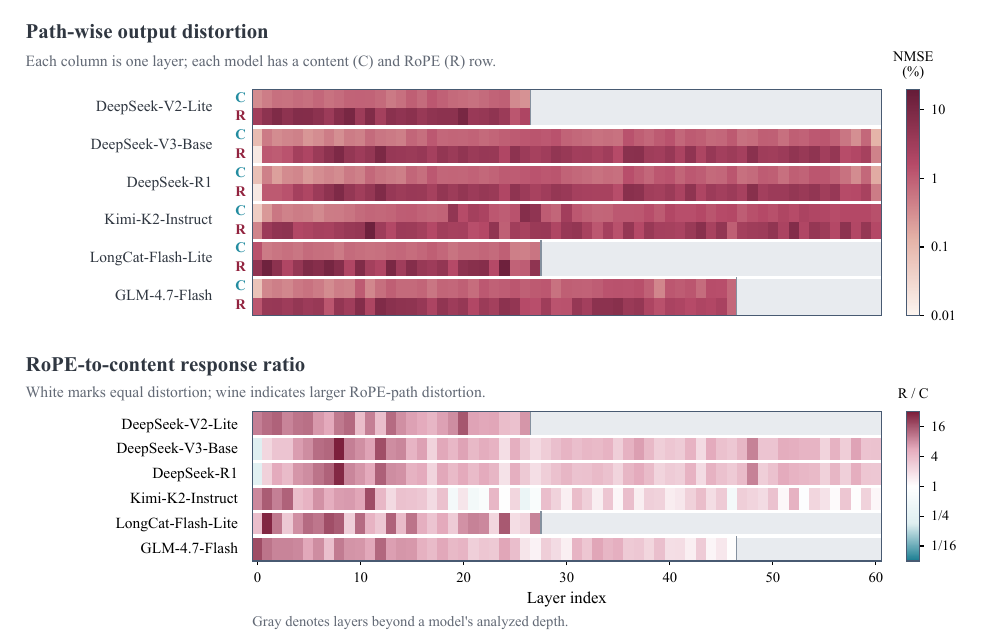}
    \caption{\textbf{Layer-wise atlas of dual-path output distortion.}
    Upper: content (C) and RoPE (R) output NMSE at cache NMSE $0.01$.
    Lower: RoPE/content ratios, with white marking equal distortion.
    Logarithmic color scales are shared across models; gray pads positions beyond each model's analyzed depth.}
    \label{fig:layerwise_error_atlas}
\end{figure}

\paragraph{Response normalization and aggregation.}
Following Section~\ref{subsec:matched_error}, the finite response of path $b$ at layer $\ell$ is
\begin{equation}
    R_{b,\ell}
    =
    \frac{\operatorname{NMSE}_{\mathrm{out},b,\ell}}
    {\operatorname{NMSE}_{\mathrm{cache},b,\ell}}.
\end{equation}
We first average each path's response over the six cells and then form the layer-wise ratio $r_\ell=\overline R_{P,\ell}/\overline R_{C,\ell}$.
The model-level ratio divides the mean RoPE response by the mean content response over layers and is therefore not the arithmetic mean of $r_\ell$.
In Figure~\ref{fig:layerwise_error_amplification}, shaded envelopes show the minimum and maximum cell responses, while diamonds and whiskers summarize the median and interquartile range of the layer-wise ratios.
Table~\ref{tab:matched_models} reports the model-level aggregates, and Figure~\ref{fig:layerwise_error_atlas} provides the complete 285-layer view.
Because the cache NMSE is fixed at $0.01$, the corresponding output NMSE is $0.01R_{b,\ell}$; its relation to local directional gain is derived in Appendix~\ref{app:gain_derivation}.

\begin{table}[ht]
\centering\small
\caption{\textbf{Model-level responses at matched cache NMSE.} Ratios use unrounded source aggregates; displayed responses are rounded.}
\label{tab:matched_models}
\begin{tabular}{lrrr}
\toprule
Model & Content response & RoPE response & Ratio \\
\midrule
DeepSeek-V2-Lite & 0.620 & 3.925 & 6.331 \\
DeepSeek-V3-Base & 0.452 & 1.751 & 3.871 \\
DeepSeek-R1 & 0.494 & 1.784 & 3.611 \\
Kimi-K2-Instruct & 1.594 & 3.095 & 1.941 \\
LongCat-Flash-Lite & 0.594 & 4.077 & 6.864 \\
GLM-4.7-Flash & 1.026 & 2.462 & 2.400 \\
\bottomrule
\end{tabular}
\end{table}

\subsection{Error Propagation and Amplification}
\label{app:gain_derivation}
\label{app:derivations}

\paragraph{Finite error routes.}
Using the absorbed computation in Equation~\ref{eq:mla_compact}, write $\widehat A_{b,h}=A_h+\Delta A_{b,h}$ for the attention probabilities after perturbing path $b$.
The content path perturbs both the attention probabilities and the shared latent, giving
\begin{align}
    \Delta Y_C
    &=\sum_h\bigl[(A_h+\Delta A_{C,h})(C+\Delta C)B_h-A_hCB_h\bigr]\\
    &=\sum_h\bigl[\Delta A_{C,h}CB_h+A_h\Delta C B_h+\Delta A_{C,h}\Delta C B_h\bigr].
\end{align}
The RoPE path leaves $C$ unchanged, so
\begin{equation}
    \Delta Y_P=\sum_h\Delta A_{P,h}CB_h.
\end{equation}
These identities hold for finite perturbations and separate the content path's matching, aggregation, and interaction terms without invoking a linear approximation.

\paragraph{Local gain and normalization.}
For a sufficiently small path-isolated quantization error $e_b$, first-order expansion gives $\Delta y_b=J_be_b+o(\|e_b\|_2)$ and hence
\begin{equation}
    \operatorname{NMSE}_{\mathrm{out},b}
    \approx
    \operatorname{NMSE}_{\mathrm{cache},b}
    \frac{\|c_b\|_2^2}{\|y\|_2^2}
    G_b(e_b).
\end{equation}
Thus, in the local regime, the finite response $R_b$ differs from the directional gain $G_b$ by the cache/output energy normalization, while finite-error measurements can additionally reflect nonlinear effects.
For $e_b\ne0$ and $T_b>0$, define
\begin{equation}
    T_b=\frac{\operatorname{tr}(H_b)}{D_b},\qquad
    A_b=\frac{\sum_i(H_b)_{ii}(e_b)_i^2}{T_b\|e_b\|_2^2},\qquad
    O_b=\frac{\sum_{i\ne j}(H_b)_{ij}(e_b)_i(e_b)_j}{\|e_b\|_2^2}.
    \label{eq:operator_error_terms}
\end{equation}
Separating the diagonal and off-diagonal contributions to $e_b^\top H_be_b$ yields the exact local identity
\begin{equation}
    G_b=T_bA_b+O_b.
\end{equation}
Here, $T_b$ measures average operator sensitivity, $A_b$ captures the allocation of quantization-error energy over sensitive coordinates, and $O_b$ captures signed cross-coordinate directional coupling.
All three quantities are defined in the declared cache coordinates.
If $e_b=0$, there is no perturbation; if $T_b=0$, positive semidefiniteness implies $H_b=0$, so the local response vanishes without requiring $A_b$ to be defined.

\paragraph{Joint-path interaction.}
\label{app:joint_path_interaction}
When both paths are quantized, the first-order output perturbation is $J_Ce_C+J_Pe_P$, with
\begin{equation}
    \|J_Ce_C+J_Pe_P\|_2^2
    =
    \|J_Ce_C\|_2^2
    +
    \|J_Pe_P\|_2^2
    +
    2\langle J_Ce_C,J_Pe_P\rangle.
\end{equation}
The interaction term need not vanish, so path-isolated learning does not assume independent error routes.
\method{} optimizes the two transformations separately according to their path-specific objectives and evaluates their composition jointly.

\subsection{Predictive Validation and Controlled Interventions}
\label{app:functional_analysis}

\paragraph{Estimating operator statistics.}
We estimate the trace and diagonal of $H_b=J_b^\top J_b$ using output-space Rademacher vector--Jacobian products.
For a Rademacher probe $r$ and $g_b=J_b^\top r$,
\begin{equation}
    \E_r\|g_b\|_2^2=\operatorname{tr}(H_b),\qquad
    \E_r[g_b\odot g_b]=\operatorname{diag}(H_b).
\end{equation}
We use 256 probes per path in the declared cache coordinates.
Sampling variation in these estimates is distinct from both off-diagonal directional coupling and nonlinear finite-error effects.

\begin{figure}[!htbp]
    \centering
    \includegraphics[width=\textwidth]{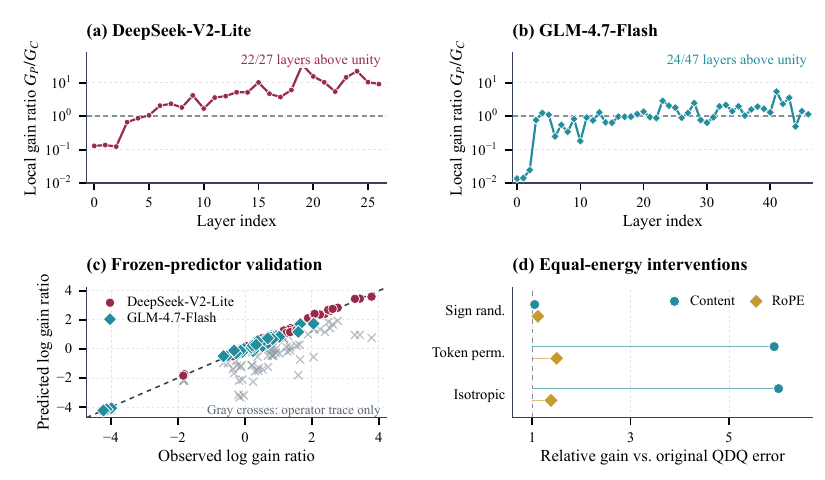}
    \caption{\textbf{Predicting and testing functional amplification.}
    Layer-wise gains, 72 validation points, and equal-energy interventions separate operator sensitivity from quantization-error geometry at cache NMSE $10^{-3}$.}
    \label{fig:functional_geometry}
\end{figure}

\paragraph{Validation protocol.}
The analysis contains 146 layer--input points in total: all 27 analyzed DeepSeek-V2-Lite layers and 47 GLM-4.7-Flash layers for the layer sweeps, together with 36 additional validation configurations per model.
QDQ-induced quantization-error directions are rescaled to cache NMSE $10^{-3}$ for this local analysis, compared with $0.01$ in the six-model finite-response study.
Figure~\ref{fig:functional_geometry} summarizes the two layer sweeps, the 72 additional validation points, and the equal-energy interventions.

\paragraph{Frozen-predictor evaluation.}
For positive gain ratios, we use
\begin{equation}
    \widehat{\log(G_P/G_C)}
    =
    \log(T_P/T_C)
    +
    \log(A_P/A_C)
    +
    \gamma,
    \qquad
    \gamma=-0.0749614410.
\end{equation}
The correction $\gamma$ is fixed from the earlier DeepSeek-V2-Lite stage and is not refitted on GLM-4.7-Flash.
It statistically accounts for effects beyond the modeled operator-sensitivity and error-allocation terms without replacing the exact decomposition in Equation~\ref{eq:gain_decomposition}.
The trace-only control retains $\log(T_P/T_C)+\gamma$, while the cache-error-only control predicts $\gamma$ under matched cache error.
Table~\ref{tab:mechanism_review} shows that incorporating quantization-error allocation substantially improves prediction over either control.

\begin{table}[!htbp]
\centering\small
\caption{\textbf{Predictive validation of the operator--error model.}
Each model contributes 36 validation points.
MAE denotes mean absolute error.}
\label{tab:mechanism_review}
\begin{tabular}{lcc}
\toprule
Model & Spearman & Median absolute log error \\
\midrule
DeepSeek-V2-Lite & 0.9923 & 0.1054 \\
GLM-4.7-Flash & 0.9614 & 0.0727 \\
\midrule
Predictor & \shortstack{DeepSeek-V2-Lite\\log-ratio MAE} & \shortstack{GLM-4.7-Flash\\log-ratio MAE} \\
\midrule
Cache error only & 1.5657 & 0.8790 \\
Operator trace only & 0.9129 & 1.4276 \\
Operator + error allocation & 0.1133 & 0.1173 \\
\bottomrule
\end{tabular}
\end{table}

\paragraph{Controlled error interventions.}
At DeepSeek-V2-Lite layer 20, WikiText/C4 and two seeds yield four configurations at cache NMSE $10^{-3}$.
The equal-energy controls independently randomize error signs, permute whole-token locations, or replace the error direction with an isotropic direction (Table~\ref{tab:error_interventions}).
Whole-token permutation increases content-path gain by $5.918\times$, close to the isotropic control, whereas sign randomization produces little change.
These results support token-dependent quantization-error allocation as a major source of suppressed content-path gain.
A JVP-versus-finite-perturbation comparison gives Spearman $0.986$ over 24 path--direction observations, supporting the local approximation at the tested error scale.

\begin{table}[!htbp]
\centering\small
\caption{\textbf{Equal-energy interventions isolate quantization-error allocation.}
Geometric-mean gain ratios relative to the original QDQ-induced errors use 32 repetitions per control across four DeepSeek-V2-Lite layer-20 configurations.
Unity indicates unchanged gain.}
\label{tab:error_interventions}
\begin{tabular}{lrr}
\toprule
Error intervention & Content & RoPE \\
\midrule
Independent sign randomization & 1.048 & 1.115 \\
Whole-token permutation & 5.918 & 1.495 \\
Isotropic direction & 6.007 & 1.380 \\
\bottomrule
\end{tabular}
\end{table}

Across layers, each model's 36-point validation set includes 16 whole-token permutations per point, yielding 576 interventions per model.
Median content-path gain increases by $2.853\times$ on DeepSeek-V2-Lite and $4.220\times$ on GLM-4.7-Flash.
These controls preserve quantization-error energy while changing its allocation, further supporting the operator--error explanation beyond the single-layer intervention.
\section{Transformation Learning and Native Execution}
\label{app:fusion}

We provide the fusion identities and positional QK error bound underlying Section~\ref{sec:quantmla}, followed by quantization, transformation learning, native execution, and correctness verification.

\subsection{Content-Path Fusion}
\label{app:content_fusion}

Write the original content latent as $c=u\Gamma$, where $u=\operatorname{RMSNorm}_1(x)$, following the row-vector convention of Section~\ref{subsec:transformation_spaces}.
Because $R_C$ is orthogonal,
\begin{equation}
    \operatorname{RMSNorm}_1(xR_C)
    =
    \operatorname{RMSNorm}_1(x)R_C.
\end{equation}
The producing projection, normalization scale, and each consuming projection can therefore be reparameterized offline as
\begin{equation}
    W_a'=W_aR_C,\qquad
    \Gamma'=S_C,\qquad
    W_b'=S_C^{-1}R_C^\top\Gamma W_b.
\end{equation}
The transformed cache $z_C=uR_CS_C$ preserves each full-precision consumer output:
\begin{equation}
    z_CW_b'=u\Gamma W_b.
\end{equation}
Thus, the learned content transformation is fully absorbed into existing model parameters and introduces no online transformation.
The update applies only to the content-latent sub-block; any concatenated positional block remains unchanged.
Producing biases, when present, undergo the same rotation, and model-specific factorizations are fused at their native projection interfaces.

\subsection{RoPE-Path Fusion}
\label{app:rope_fusion}

\paragraph{Equivalent positional reparameterization.}
Let $R_P=\operatorname{blockdiag}(R_i)$ with each $R_i\in SO(2)$ acting on one RoPE frequency pair, and let $S_P=\operatorname{blockdiag}(s_iI_2)$ with $s_i>0$.
The reciprocal query--key transformation in Equation~\ref{eq:rope_reciprocal_transform} preserves full-precision positional scores and commutes with every positional rotation under the same pair layout, permitting fusion into the pre-RoPE projections.

\paragraph{Proof.}
Within each frequency pair, both $R_i$ and the positional rotation are planar rotations and therefore commute; the scalar matrix $s_iI_2$ commutes with both.
The block-diagonal transforms inherit these relations under the same native pair layout.
Moreover, $R_PR_P^\top=I$ and $S_P^{-1}S_P^\top=I$, so
\begin{equation}
    Q_P'(K_P')^\top
    =
    Q_PR_PS_P^{-1}S_P^\top R_P^\top K_P^\top
    =
    Q_PK_P^\top.
\end{equation}
Thus, the learned transformation and reciprocal query compensation can be folded into the pre-RoPE projections while preserving the positional scores.
For a row-vector projection $hW$,
\begin{equation}
    W_K'=W_KR_PS_P,\qquad
    W_Q'=W_QR_PS_P^{-1}.
\end{equation}
The same updates apply to relevant biases and only to the positional sub-block when a projection produces additional features.
The implementation follows each model's native RoPE pairing convention and introduces no cross-frequency online mixing.

\subsection{Bounding Output Error by Positional QK Error}
\label{app:qk_bound}

We prove Proposition~\ref{prop:qk_output_bound} and derive its normalized extension below.
Hold the content logits and values fixed, write the positional score contribution as $Z_P=Q_PK_P^\top$, and let $\Omega$ select valid causal query--key pairs.
Assume every query has at least one valid key and that softmax is evaluated over the same valid-key set before and after perturbation.
For a probability vector $a$, the softmax Jacobian
\begin{equation}
    D=\operatorname{Diag}(a)-aa^\top
\end{equation}
is symmetric positive semidefinite.
The absolute sum of row $i$ is $2a_i(1-a_i)\leq1/2$, giving
\begin{equation}
    \|D\|_2
    \leq
    \sqrt{\|D\|_1\|D\|_\infty}
    \leq
    \frac{1}{2}.
\end{equation}
Integrating the Jacobian along the segment between the original and perturbed score vectors yields
\begin{equation}
    \|\widehat A-A\|_F
    \leq
    \frac{\tau}{2}
    \|\Omega\odot(\widehat Z_P-Z_P)\|_F.
\end{equation}
For one head, let $\bar V_h=CB_h=V_hW_{O,h}$ denote the effective projected values.
Then
\begin{equation}
    \|\widehat Y_h-Y_h\|_F
    \leq
    \frac{\tau}{2}
    \|\Omega\odot(\widehat Z_{P,h}-Z_{P,h})\|_F
    \|\bar V_h\|_2.
\end{equation}
With shared positional keys and multiple heads, the layer output perturbation is the sum of projected head perturbations.
Applying the triangle inequality and Cauchy--Schwarz gives
\begin{equation}
    \|\Delta Y_P\|_F
    \leq
    \frac{\tau}{2}
    \left(\sum_h\|\Omega\odot\Delta Z_{P,h}\|_F^2\right)^{1/2}
    \left(\sum_h\|\bar V_h\|_2^2\right)^{1/2}.
\end{equation}
Squaring both sides gives Equation~\ref{eq:qk_output_bound}.
Unlike the directional-gain analysis, this bound applies to finite positional-score perturbations and requires no first-order approximation.
It assumes fixed queries, content scores, values, and valid-key sets while varying the RoPE cache; quantizing the content path introduces additional terms.
For any fixed candidate transformation, equivalent query--key compensation preserves the full-precision positional scores, so the bound applies pointwise to every candidate representation even though the compensation changes during learning.

\paragraph{Normalized objective.}
For a fixed calibration sample and layer, define
\begin{equation}
    D_Z=\max\!\left\{\sum_h\|\Omega\odot Z_{P,h}\|_F^2,\epsilon_P\right\},
    \qquad
    D_Y=\max\{\|Y\|_F^2,\epsilon_C\}.
\end{equation}
Let
\begin{equation}
    \ell_P
    =
    \frac{\sum_h\|\Omega\odot\Delta Z_{P,h}\|_F^2}{D_Z}
\end{equation}
be the sample-level loss corresponding to Equation~\ref{eq:rope_qk_objective}.
Then
\begin{equation}
    \frac{\|\Delta Y_P\|_F^2}{D_Y}
    \leq
    \frac{\tau^2D_Z}{4D_Y}
    \left(\sum_h\|\bar V_h\|_2^2\right)\ell_P.
\end{equation}
For a fixed sample, the multiplier is independent of the learned RoPE transformation because equivalent compensation preserves teacher positional scores while values remain fixed, although it may vary across samples and layers.
Thus, mean normalized positional QK reconstruction is an output-controlling surrogate rather than an exact reformulation of mean output reconstruction.
On a finite calibration set, the largest multiplier bounds the mean output error by the mean QK loss.
Retaining sample-specific multipliers tightens the bound but changes the learning objective.

\paragraph{Functional specificity of positional QK reconstruction.}
For a valid key $k$ and query row $q$,
\begin{equation}
    \frac{\partial Y_{h,q}}{\partial (Z_{P,h})_{qk}}
    =
    \tau A_{h,qk}\bigl(V_{h,k}-O_{h,q}\bigr)W_{O,h}.
\end{equation}
Attention-output reconstruction and positional QK reconstruction therefore need not weight the same score errors equally.
The former weights perturbations through attention probabilities, values, and the output projection, whereas the latter directly preserves positional matching.
Positional QK reconstruction therefore preserves the RoPE-induced component of the attention logits, with output distortion controlled by the bound above.

\subsection{Quantization and Transformation Learning}
\label{app:learning}
\label{app:calibration}
\label{app:algorithm}

\paragraph{Affine group quantization.}
Each token is quantized independently in groups of 64 coordinates.
For bit width $b$, let $m=2^b-1$ and let $x_{\min},x_{\max}$ be the group extrema.
The reference quantizer uses
\begin{align}
    s&=\frac{\max(x_{\max}-x_{\min},10^{-8})}{m},
    & z&=\operatorname{clip}\!\left(\operatorname{round}(-x_{\min}/s),0,m\right),\\
    q&=\operatorname{clip}\!\left(\operatorname{round}(x/s+z),0,m\right),
    & \widehat x&=s(q-z).
\end{align}
We use round-to-nearest with ties to even, without special handling of constant groups.
Calibration applies a straight-through estimator to rounding.
Calibration and native execution share the same scale-computation, zero-point, rounding, and clipping rules.
Native execution stores content scales in BF16 and RoPE scales in FP32 and dequantizes using the stored scales.

\paragraph{Function-aligned optimization.}
All pretrained parameters remain frozen during learning.
Content rotations use a Cayley parameterization, RoPE-pair rotations use trainable angles, and positive scales are optimized in log space.
The content path minimizes attention-output reconstruction, while the RoPE path minimizes positional QK reconstruction over all valid causal pairs and heads.
The latter uses the complete positional QK matrix rather than separate frequency-pair errors, preserving cross-frequency cancellation.
Calibration has no BF16-protected region; the sink/local policy is applied only at evaluation.
Each path selects the checkpoint with the lowest held-out path objective, without downstream test selection.
Table~\ref{tab:calibration} specifies the optimization settings.

\begin{table}[!htbp]
\centering\small
\caption{\textbf{Calibration and transformation-learning settings.}
Learning uses disjoint training and validation banks; downstream comparisons use seed 0.}
\label{tab:calibration}
\begin{tabular}{ll}
\toprule
Item & Setting \\
\midrule
Calibration corpus & WikiText-2 training split \\
Sequence length / batch size & 2048 / 4 \\
Train / validation bank & 128 / 32 sequences per seed \\
Optimization & 300 steps per path \\
Optimizer & AdamW, weight decay 0 \\
Schedule & 10\% warmup, cosine decay \\
Gradient clipping & 1.0 \\
Content rotation LR & $3\times10^{-3}$ \\
RoPE angle / scale LR & $1\times10^{-3}$ / $1\times10^{-3}$ \\
Content scale LR & $1\times10^{-3}$ \\
Rotation / angle / scale anchor & $10^{-4}$ \\
Validation frequency & Every 20 steps, including step 0 \\
Calibration seeds & 0, 1, 2 \\
Quantizer & Affine asymmetric, group size 64 \\
Scale-initialization search & $\alpha\in\{0,.125,.25,.5,.75,1\}$ \\
\bottomrule
\end{tabular}
\end{table}

The content loss normalizes mean squared error by teacher mean-square energy clamped at $10^{-8}$; equivalently, $\epsilon_C=N_Y10^{-8}$ for $N_Y$ output entries in Equation~\ref{eq:content_objective}.
The positional loss uses summed squared error and teacher positional-score energy clamped at $\epsilon_P=10^{-8}$.

\paragraph{Scale initialization.}
Let $a_j$ denote the $99.9$th-percentile magnitude of channel $j$, and define $\widetilde a_j=\max(a_j,10^{-8})$.
We initialize
\begin{equation}
    \log s_j^{(0)}
    =
    \operatorname{clip}\!\left(
    \alpha\left[\operatorname{mean}_k\log \widetilde a_k-\log \widetilde a_j\right],-2,2\right).
    \label{eq:scale_initialization}
\end{equation}
During learning and offline fusion, scales are obtained by exponentiating log-scale parameters clipped to $[-2,2]$.
RoPE statistics and scales are tied within each frequency pair.
The held-out search includes $\alpha=0$ and favors zero in a tie.
Content rotations start from identity in weightless-normalized coordinates, which need not reproduce native-cache RTN on $u\Gamma$.

\paragraph{Calibration procedure.}
We capture BF16 layer inputs, attention outputs, and positional queries and keys, then initialize the two transformations using the held-out scale search.
For each path, we optimize its transformation against the corresponding function-aligned objective while keeping the other path exact; the best checkpoint is selected independently.
After checking precision, quantizer, and checkpoint metadata, we compose the two learned transformations and fuse their parameters into the producing and consuming projections.
Quantization-disabled equivalence is verified before evaluating the jointly quantized model.
No learned transformation or fixed mixing operator remains online.

\subsection{Native Cache Layout and Mixed-Precision Execution}
\label{app:kernel}

\begin{figure}[!htbp]
    \centering
    \includegraphics[width=\textwidth]{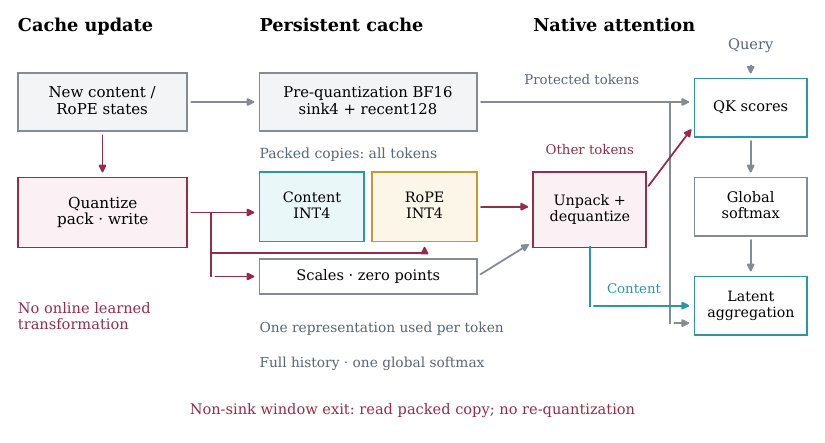}
    \caption{\textbf{Native mixed-precision MLA execution.}
    The fused writer generates a packed C4R4 copy for every incoming token and retains the pre-quantization BF16 values of protected sink and recent tokens in dedicated buffers.
    Attention uses pre-quantization BF16 values for protected tokens and dequantized INT4 values otherwise.
    When a non-sink token leaves the recent window, execution switches to its existing packed copy without additional quantization.
    Reconstructed content tiles support both matching and latent aggregation, while all valid tokens share one global softmax normalization.}
    \label{fig:kernel_dataflow}
\end{figure}

The backend directly stores the shared content latent and decoupled RoPE key in the coordinates defined by the offline-fused transformations.
Figure~\ref{fig:kernel_dataflow} summarizes the physical cache layout and read path without per-head KV expansion.

\paragraph{Cache update and tile reuse.}
The fused writer performs content- and RoPE-cache quantization, integer packing, slot-mapped page insertion, and updates to the BF16 protection buffers.
A packed copy is stored for every token, including tokens whose pre-quantization BF16 representation is currently protected.
During decoding, the attention kernel gathers cache pages, unpacks integer codes, and reconstructs content and RoPE tiles using the stored scales and zero points.
Each reconstructed content tile is reused for both content matching and latent aggregation and remains available until both consumers complete.
Query preparation, split-result combination, and value/output projection retain their native execution boundaries.

\paragraph{Shared-memory organization and pipelining.}
In the optimized decode kernel, packed inputs and quantization metadata are buffered separately from reconstructed BF16 tiles, allowing prefetch to proceed while earlier tiles remain in use.
Independent metadata storage removes aliasing with reconstructed RoPE tiles, enabling earlier RoPE QK scheduling without changing the accumulation order.
Protected BF16 content loads overlap with INT4 reconstruction of unprotected rows, with synchronization before consumption.
Packed buffers are reused after reconstruction completes, whereas reconstructed content remains available through both QK and PV and is overwritten only after its consumers finish.

\paragraph{BF16 protection and cache lifecycle.}
For $L$ cached tokens with zero-based indices, the protected set is
\begin{equation}
    \mathcal P_L
    =
    \{0,\ldots,\min(4,L)-1\}
    \cup
    \{\max(0,L-128),\ldots,L-1\}.
\end{equation}
The union counts overlapping sink and recent regions only once.
For tokens in $\mathcal P_L$, attention uses the pre-quantization BF16 content and RoPE-key values; all remaining tokens use values reconstructed from the packed cache.
Each token therefore contributes exactly once to attention, although protected tokens retain both BF16 and packed representations in storage.
When a non-sink token leaves the recent window, attention switches directly to the packed copy generated at insertion, requiring no additional quantization or transfer.
This policy changes storage precision rather than attention visibility: no historical token is evicted or masked.

\paragraph{Global attention normalization.}
Protected and quantized regions participate in the same softmax normalization.
If the regions are processed separately, let $m_j$, $s_j$, and $o_j$ denote their partial maximum, exponential sum, and unnormalized weighted-value sum.
They are merged as
\begin{equation}
    m=\max_j m_j,\qquad
    s=\sum_j e^{m_j-m}s_j,\qquad
    o=\sum_j e^{m_j-m}o_j,\qquad
    O=o/s.
\end{equation}
This online merge preserves full-history attention normalization; independently normalizing region outputs would not.

\paragraph{Storage layout.}
The persistent cache contains packed copies of all tokens, quantization metadata, page alignment, fixed BF16 protection buffers, and reverse page mappings.
For $d_c=512$, $d_r=64$, and 64-token pages, the packed C4R4 representation occupies $20{,}480$ bytes per page, or $320$ bytes per token and layer.
Content scales are stored in BF16 and RoPE scales in FP32.
Each request and layer additionally allocates
$132\times576\times2=152{,}064$ bytes for BF16 protection, while reverse page mappings require 8 bytes per physical page.
The corresponding BF16 cache page occupies
$64\times1{,}152=73{,}728$ bytes.
For a request of length $L$, the per-layer persistent allocations are
\begin{equation}
\begin{aligned}
M_{\mathrm{C4R4}}(L)&=\left\lceil L/64\right\rceil(20{,}480+8)+152{,}064,\\
M_{\mathrm{BF16}}(L)&=\left\lceil L/64\right\rceil73{,}728.
\end{aligned}
\label{eq:physical_cache_storage}
\end{equation}
Their ratio is $3.22\times$ at 4K and $3.59\times$ at 128K.
These allocations include redundant packed copies of protected tokens but exclude model weights, shared page tables, allocator overhead, and temporary workspaces.

\paragraph{Fusion and execution verification.}
We first disable quantization and compare the original and offline-fused models using layer-output error and end-to-end NLL drift.
This isolates algebraic fusion from quantization fidelity and detects errors in RoPE pairing, transformation order, sub-block selection, or reciprocal compensation.
Native execution is checked against a BF16 reference consuming the same independently reconstructed cache, including dynamic cache updates and token aging.
These checks validate cache representation and execution semantics rather than agreement with the original unquantized model.
Appendix~\ref{app:system_efficiency} summarizes the validation scope; shared-prefix ownership remains outside it.

\section{Evaluation Protocols}
\label{app:reproducibility}

This appendix specifies task evaluation, the MLA-specific baseline adaptations, and the component-ablation protocol.
Quantization and transformation learning for QuantMLA are detailed in Appendix~\ref{app:calibration}.

\subsection{Task Evaluation and Scoring}
\label{app:task_protocol}

\paragraph{Knowledge and reasoning.}
MMLU~\citep{mmlu} uses 5-shot evaluation over all 14,042 examples.
GSM8K~\citep{gsm8k} uses 5-shot generation and exact match after strict answer extraction.
MATH500~\citep{math} uses 4-shot prompting with Math-Verify accuracy, while GPQA-Diamond~\citep{gpqa} uses zero-shot prompting and accuracy over all 198 four-choice questions.
AIME25 uses all 30 problems with zero-shot chat prompting and exact final-answer matching.
The same prompting and scoring protocol is applied across methods within each benchmark.

\paragraph{Code generation.}
HumanEval~\citep{humaneval} reports single-sample pass@1.
LiveCodeBench~\citep{livecodebench} evaluates all 1,055 problems in \texttt{code\_generation\_lite}, \texttt{release\_v6}, with one deterministic candidate per problem.
Generation uses temperature zero, a 2,000-token output budget, and thinking disabled; we report benchmark pass@1.

\paragraph{Commonsense and information extraction.}
HellaSwag, PIQA, ARC-Easy, and ARC-Challenge use length-normalized accuracy (\texttt{acc\_norm}), while WinoGrande uses accuracy (\texttt{acc})~\citep{hellaswag,piqa,arc,winogrande}.
The FDA, SWDE, and SQuAD-Completion variants distributed with Based~\citep{based} contain 1,102, 1,111, and 2,984 validation examples, respectively.
SQuAD-Completion reformulates SQuAD~\citep{squad} as a completion task and is labeled SQuAD in our tables.
All three tasks use zero-shot generation and case-insensitive answer containment, with either a newline stop or a 48-token output limit.
The corresponding datasets are \texttt{hazyresearch/based-fda}, \texttt{hazyresearch/based-swde-v2}, and \texttt{hazyresearch/based-squad}.

\paragraph{Aggregation and presentation.}
Suite averages assign equal weight to the benchmarks within each reported suite.
Bold entries in the main tables denote the best quantized result within each model, precision, and metric, including ties at the displayed precision; BF16 is excluded from this comparison.

\subsection{Baseline Adaptations}
\label{app:baseline_scope}

SmoothQuant$^\dagger$ and QuaRot$^\dagger$ are our adaptations of SmoothQuant~\citep{smoothquant} and QuaRot~\citep{quarot} to MLA KV-cache quantization, rather than direct reproductions of their original full-model quantization pipelines.
Both operate directly on MLA's native content latent and decoupled RoPE key cache without expanding them into per-head keys and values.
For accuracy evaluation, both use the same per-token asymmetric affine quantizer with group size 64 as QuantMLA and are evaluated through cache quantize--dequantize simulation.

\paragraph{SmoothQuant$^\dagger$.}
This adaptation retains the equivalent channel-scaling principle of SmoothQuant while adapting it to MLA's two cache components.
For the content cache, let $u$ denote the weightless-normalized latent and $\Gamma$ the original RMSNorm channel gain.
We cache the rescaled representation $uS_C$ and compensate all downstream content consumers, so that both the key and value routes preserve the original full-precision computation when quantization is disabled.
For the RoPE key cache, the two coordinates within each RoPE frequency pair share one positive scale, with reciprocal query--key compensation,
\begin{equation}
    K'_P = K_P S_P,
    \qquad
    Q'_P = Q_P S_P^{-1},
\end{equation}
which preserves the positional QK scores in full precision.
All scaling and compensation are absorbed into existing normalization and projection parameters, introducing no online scaling operation.

The scales are determined without gradient optimization.
For channel magnitude statistic $a_j$, we define $\widetilde a_j=\max(a_j,10^{-8})$ and construct candidates as
\begin{equation}
    \log s_j
    =
    \operatorname{clip}
    \left(
        \alpha
        \left[
            \operatorname{mean}_k \log \widetilde a_k
            -
            \log \widetilde a_j
        \right],
        -2,2
    \right),
\end{equation}
where $a_j$ is the $99.9$th-percentile absolute activation magnitude.
RoPE statistics and scales are tied within each frequency pair.
Candidate values of $\alpha$ are drawn from $\{0.125,0.25,0.5,0.75,1\}$ according to the recorded configuration.
Statistics are collected from 128 WikiText-2 sequences of length 2048, and candidates are selected on a disjoint set of 32 sequences.
The content and RoPE caches are calibrated separately, keeping the other cache exact, and both select $\alpha$ by held-out normalized attention-output MSE.
Thus, SmoothQuant$^\dagger$ uses statistical scaling with validation-based selection rather than the learned path-specific objectives of QuantMLA.

\paragraph{QuaRot$^\dagger$.}
This adaptation retains the fixed Hadamard-rotation principle of QuaRot without data-driven rotation learning.
For the content cache, a deterministic normalized Hadamard matrix $H_C$ is applied in the weightless-normalized latent coordinates.
The rotation is fused into the producing projection using RMSNorm's orthogonal equivariance, while the inverse transformation and original normalization gain are absorbed into the key and value consumers.
The content-cache rotation is therefore entirely offline and introduces no per-token Hadamard operation during inference.

For the RoPE cache in the INT4 configurations used in our main experiments, the same fixed normalized Hadamard matrix $H_P$ is applied to the positional query and key after RoPE:
\begin{equation}
    Q'_P = Q_P H_P,
    \qquad
    K'_P = K_P H_P.
\end{equation}
Since $H_PH_P^\top=I$, this transformation preserves positional QK scores before quantization.
Only the transformed key is cached and quantized, while the corresponding query transformation is evaluated online.
Unlike QuantMLA's RoPE-compatible offline transformation, this post-RoPE Hadamard cannot be fused into the pre-RoPE projections and therefore remains an online operation.
QuaRot$^\dagger$ uses deterministic normalized Hadamard matrices without learned rotations, random sign matrices, statistical calibration, or gradient optimization.

\subsection{Component-Ablation Protocol}
\label{app:component_protocol}

Table~\ref{tab:ablations} evaluates the full DeepSeek-V2-Lite MMLU test set at C4R4.
Starting from unprotected RTN, we successively add the content-path transformation, the RoPE-path transformation, and the mixed-precision cache policy.
The two transformation increments are therefore measured without BF16-protected tokens, while the final configuration retains four sink tokens and the most recent 128 tokens in BF16.
The resulting sequence is $51.90\% \rightarrow 53.12\% \rightarrow 56.31\% \rightarrow 57.95\%$, compared with $57.90\%$ for BF16.
The SmoothQuant$^\dagger$ and QuaRot$^\dagger$ rows are independent unprotected baselines and are not part of this cumulative sequence.
The reported component gains are cumulative along the stated order and should not be interpreted as order-independent effects.
\section{Additional Accuracy Results}
\label{app:full_results}

Table~\ref{tab:commonsense_details} disaggregates the Commonsense averages reported in Table~\ref{tab:general_results} into the five constituent benchmarks for DeepSeek-V2-Lite and Moonlight-16B-A3B.
We provide this breakdown because the remaining task-level metrics are already reported individually in the main tables.

\begin{table}[!htbp]
\centering
\small
\caption{\textbf{Detailed commonsense performance.}
Scores (\%, $\uparrow$) on HellaSwag, PIQA, ARC-Easy, ARC-Challenge, and WinoGrande.
$\dagger$: our MLA adaptations.}
\label{tab:commonsense_details}
\setlength{\tabcolsep}{3.2pt}
\begin{tabular}{@{}llrrrrrr@{}}
\toprule
Precision & Method
& HellaSwag
& PIQA
& \shortstack{ARC-\\Easy}
& \shortstack{ARC-\\Challenge}
& \shortstack{Wino-\\Grande}
& Avg. \\
\midrule

\multicolumn{8}{c}{\textit{DeepSeek-V2-Lite}} \\
\midrule
BF16 & BF16
& 77.76
& 79.87
& 74.28
& 46.16
& 70.72
& 69.76 \\
\addlinespace[2pt]
C4R4 & RTN
& 75.20
& 78.45
& 70.03
& 44.71
& 67.48
& 67.17 \\
& SmoothQuant$^\dagger$
& 75.64
& 78.73
& 71.97
& 44.11
& 70.72
& 68.23 \\
& QuaRot$^\dagger$
& 76.64
& 79.27
& 73.86
& 45.73
& 69.53
& 69.01 \\
& \method{}
& 77.84
& 80.14
& 74.33
& 46.50
& 70.80
& 69.92 \\\midrule
\addlinespace[2pt]
C2R4 & RTN
& 69.69
& 75.41
& 66.16
& 38.14
& 61.64
& 62.21 \\
& SmoothQuant$^\dagger$
& 73.36
& 77.58
& 69.82
& 42.58
& 66.30
& 65.93 \\
& QuaRot$^\dagger$
& 72.35
& 75.57
& 67.72
& 41.55
& 64.40
& 64.32 \\
& \method{}
& 77.82
& 80.36
& 74.16
& 46.42
& 70.48
& 69.85 \\

\midrule

\multicolumn{8}{c}{\textit{Moonlight-16B-A3B}} \\
\midrule
BF16 & BF16
& 78.28
& 80.79
& 82.53
& 58.02
& 71.98
& 74.32 \\
\addlinespace[2pt]
C4R4 & RTN
& 77.78
& 80.90
& 83.29
& 57.25
& 70.48
& 73.94 \\
& SmoothQuant$^\dagger$
& 77.26
& 81.01
& 81.99
& 55.97
& 70.72
& 73.39 \\
& QuaRot$^\dagger$
& 78.59
& 80.63
& 81.82
& 57.68
& 71.82
& 74.11 \\
& \method{}
& 78.34
& 80.58
& 82.58
& 58.28
& 71.43
& 74.24 \\\midrule
\addlinespace[2pt]
C2R4 & RTN
& 74.21
& 77.97
& 77.99
& 50.68
& 68.03
& 69.78 \\
& SmoothQuant$^\dagger$
& 70.69
& 75.90
& 68.43
& 45.31
& 64.40
& 64.95 \\
& QuaRot$^\dagger$
& 67.37
& 75.52
& 67.63
& 42.24
& 60.85
& 62.72 \\
& \method{}
& 78.32
& 80.85
& 82.41
& 57.94
& 71.59
& 74.22 \\
\bottomrule
\end{tabular}
\end{table}
\section{Additional Efficiency Results}
\label{app:efficiency}

\subsection{Native Attention}
\label{app:system_protocol}
\label{app:system_efficiency}
We measure C4R4 with the mixed-precision cache policy (sink4/local128) on one GPU, using 16 query heads, one query token, QK dimension 576, latent aggregation dimension 512, 64-token pages, and 78 scheduling partitions.
Inputs are synthetic BF16 queries and cache values with independently owned physical pages.
All shapes use the same C4R4 backend, with context length and maximum-length hint both set to $N$.
Each configuration is measured in two independent runs, each with two rounds in reversed backend order and 30 samples per round, yielding 120 samples per configuration and backend; we average both runs.
CUDA Graph replay executes cache writing, metadata construction, attention-main, and split-result combination, with L2 flushed before each graph sample.
The reported latency spans attention-main start through combination end, including unpacking and dequantization but excluding the preceding writer and metadata operations.
The performance baseline consumes an unquantized BF16 cache with native partitioning; the comparison reflects both cache representation and kernel scheduling.
Table~\ref{tab:system_latency} and Figure~\ref{fig:efficiency}(b) report paired results for batches 8 and 32 at fixed context lengths from 1K to 1M.

Correctness is checked separately against a BF16 reference consuming the same reconstructed mixed-precision cache with the same partitioning.
Output and log-sum-exp agree bit for bit for every reported configuration in both runs.
These checks validate native execution rather than equivalence to unquantized attention.
Physical storage accounting is given in Appendix~\ref{app:kernel}.

\begin{table}[!htbp]
\centering\small
\setlength{\tabcolsep}{6pt}
\caption{\textbf{Attention latency at fixed context lengths on GPU ($\mu$s).} C4R4 uses the mixed-precision cache policy (sink4/local128). Each entry averages 120 samples from two independent runs; $\Delta=(t_{\mathrm{C4R4}}/t_{\mathrm{BF16}}-1)\times100\%$. BF16 uses native partitioning. Here $1\mathrm{K}=1024$ and $1\mathrm{M}=1{,}048{,}576$ tokens.}
\label{tab:system_latency}
\begin{tabular}{@{}rrrrr@{}}
\toprule
Batch & Context $N$ & BF16 & C4R4 & $\Delta$ (\%) \\
\midrule
8 & 1K & 25.69 & 23.78 & -7.42 \\
8 & 4K & 51.43 & 53.63 & +4.29 \\
8 & 8K & 82.50 & 85.35 & +3.45 \\
8 & 16K & 147.34 & 149.79 & +1.66 \\
8 & 32K & 281.44 & 281.99 & +0.20 \\
8 & 64K & 542.14 & 540.61 & -0.28 \\
8 & 128K & 1067.24 & 1062.32 & -0.46 \\
8 & 256K & 2122.10 & 2113.33 & -0.41 \\
8 & 512K & 4234.79 & 4214.84 & -0.47 \\
8 & 1M & 8440.05 & 8418.63 & -0.25 \\
\midrule
32 & 1K & 55.50 & 56.97 & +2.64 \\
32 & 4K & 156.82 & 155.54 & -0.82 \\
32 & 8K & 287.58 & 283.46 & -1.43 \\
32 & 16K & 548.73 & 547.20 & -0.28 \\
32 & 32K & 1072.93 & 1067.28 & -0.53 \\
32 & 64K & 2133.71 & 2124.10 & -0.45 \\
32 & 128K & 4236.96 & 4225.35 & -0.27 \\
32 & 256K & 8449.30 & 8426.35 & -0.27 \\
32 & 512K & 16880.19 & 16827.11 & -0.31 \\
32 & 1M & 33701.48 & 33609.17 & -0.27 \\
\bottomrule
\end{tabular}
\end{table}

\subsection{Serving under Cache Pressure}
\label{app:serving}
We run DeepSeek-R1-0528 on eight GPUs with attention TP8/DP1 and MoE TP8, using FP8 model weights and BF16 activations.
Both configurations use vLLM 0.10.2, PyTorch 2.8.0, a 90\% GPU-memory utilization setting, 64-token pages, disabled prefix caching, and at most 16 concurrent sequences.
Chunked prefill executes eagerly with an 8,192-token step budget; decode uses CUDA Graph replay for batches 1--16.
The KV cache is either BF16 or C4R4 with sink4/local128 protection.

Five requests arrive simultaneously and enter the native vLLM scheduler without frontend admission control.
Each uses 131,072 input tokens and generates exactly 1,024 output tokens, with temperature zero and EOS ignored.
The prompts fit within BF16 capacity, but the full input-plus-output budget of 660,480 tokens exceeds its 657,856 usable slots, causing repeated preemption and recomputation.
The C4R4 cache accommodates the workload without preemption.

Whole-job output throughput is 5,120 generated tokens divided by elapsed time, including prefill, waiting, scheduling, and recomputation, but excluding model loading and warmup.
Table~\ref{tab:serving_sweep} compares one C4R4 run with the recorded BF16 baseline; no confidence interval across independent runs is reported.
The $5.168\times$ gain reflects avoiding capacity-driven recomputation in this workload and is not an attention-kernel speedup.

\begin{table}[!htbp]
\centering\small
\caption{\textbf{Serving under native vLLM scheduling on eight GPUs.} Five simultaneous requests each use 128K input and 1K output tokens. C4R4 includes sink4/local128 protection.}
\label{tab:serving_sweep}
\begin{tabular}{@{}lrr@{}}
\toprule
Metric & BF16 KV & C4R4 KV \\
\midrule
Usable cache slots & 657,856 & 2,360,832 \\
Elapsed time (s) & 1,958.392 & 378.929 \\
Output throughput (tokens/s) & 2.614 & 13.512 \\
Preemption events & 32 & 0 \\
Recomputed tokens & 3,996,255 & 0 \\
\bottomrule
\end{tabular}
\end{table}

\end{document}